\documentclass[letterpaper, 10 pt, conference]{ieeeconf}
\IEEEoverridecommandlockouts
\usepackage{epsfig}
\usepackage{amssymb}
\usepackage{cite}
\usepackage{booktabs}
\usepackage{multirow}
\usepackage{mathtools}
\usepackage{bm}
\usepackage{algorithm}
\usepackage{comment}
\usepackage{color}
\usepackage{setspace}

\newcommand{\figlab}[1]{\label{fig:#1}}
\newcommand{\figref}[1]{Fig.~\ref{fig:#1}} 
\newcommand{\tablab}[1]{\label{tab:#1}}
\newcommand{\tabref}[1]{Table~\ref{tab:#1}} 

\newcommand{\etal}{\textit{et al.}}
\newcommand{\ie}{\textit{i.e.,}}
\newcommand{\eg}{\textit{e.g.,}}
\definecolor{green}{rgb}{0.01, 0.5, 0.01}
\usepackage{pifont}
\newcommand{\cmark}{\ding{51}}%
\newcommand{\xmark}{\ding{55}}%

\usepackage{subcaption}
\usepackage{url}
\usepackage{threeparttable}

\usepackage{amsmath}
\usepackage[noend]{algorithmic}
\usepackage{array}

\begin{document}
\title{Efficiently Collecting Training Dataset for 2D Object Detection \\by Online Visual Feedback}
\author{Takuya Kiyokawa$^{*}$, Naoki Shirakura, Hiroki Katayama, Keita Tomochika, Jun Takamatsu
\thanks{T. Kiyokawa is with the Department of Systems Innovation, Graduate School of Engineering Science, Osaka University, Toyonaka, Osaka, Japan. $^{*}$Corresponding author: {\tt\small kiyokawa@sys.es.osaka-u.ac.jp}}%
\thanks{N. Shirakura is with the National Institute of Advanced Industrial Science and Technology, Aomi, Koto-ku, Tokyo, Japan.}%
\thanks{H. Katayama, K. Tomochika, and J. Takamatsu are with the Division of Information Science, Nara Institute of Science and Technology (NAIST), Ikoma, Nara, Japan.}%
}

\maketitle

\vspace{-20pt}
\begin{abstract}
Training deep-learning-based vision systems require the manual annotation of a significant number of images. Such manual annotation is highly time-consuming and labor-intensive. Although previous studies have attempted to eliminate the effort required for annotation, the effort required for image collection was retained. To address this, we propose a human-in-the-loop dataset collection method that uses a web application. To counterbalance the workload and performance by encouraging the collection of multi-view object image datasets in an enjoyable manner, thereby amplifying motivation, we propose three types of online visual feedback features to track the progress of the collection status.
Our experiments thoroughly investigated the impact of each feature on collection performance and quality of operation. The results suggested the feasibility of annotation and object detection.
\end{abstract}

\IEEEpeerreviewmaketitle

\section{Introduction}
In recent years, highly accurate object detection systems with deep learning (DL) have garnered attention in the fields of logistics~\cite{Fujita2020}, waste treatment~\cite{KiyokawaTASE2022}, manufacturing~\cite{Zhang2022}, and retailing~\cite{Naphade2022}.
Their industries, where new target objects arrive quickly one after another, require the prompt collection of a new training dataset and retraining of an object detection model.
However, because it takes considerable time and effort to collect the training dataset, it is extremely difficult to complete the process from the introduction of new objects to the execution of object detection.
This problem is urgent not only for camera-only applications such as product detection, waste sorter, defective mechanical part detection, automatic cash registers, and home robots but also for automatic operations with vision-based robot systems in their fields~\cite{KiyokawaRAL2019,Suchi2019,KiyokawaAR2019,Gregorio2020,Ishida2020,Uygur2024}.

In collecting training datasets, normally with an annotation tool, we must manually draw two-dimensional (2D) bounding boxes around objects in the images and carefully label each object enclosed by each 2D bounding box. 
This annotation process is time-consuming and labor-intensive.
Noisy annotations degrade the performance of the trained models~\cite{Agnew2023}.
Manual annotation of existing datasets has been considered problematic due to its low quality (e.g., low consistency)~\cite{Murrugarra2022}, and there have been attempts to address this issue through more laborious re-annotation with guidelines~\cite{Ma2022}.
The manual annotation process presents certain subjectiveness (particularly when occlusions are present), and the quality of annotation might get degraded as annotators become fatigued.

Crowdsourcing is an efficient and cost-effective method of collecting annotated image datasets.
To control the quality of crowdsourced annotations, human-in-the-loop verification techniques~\cite{Hao2012}, web-based image annotation interfaces~\cite{Russell2008,Benenson2019}, and web-based video annotation platforms~\cite{Vondrick2013,Dutta2019} have been proposed.
However, there is always a possibility of human error, and as a result, the trained DL model did not achieve the best performance.

\begin{figure*}[tb]
  \centering
  \includegraphics[width=\linewidth]{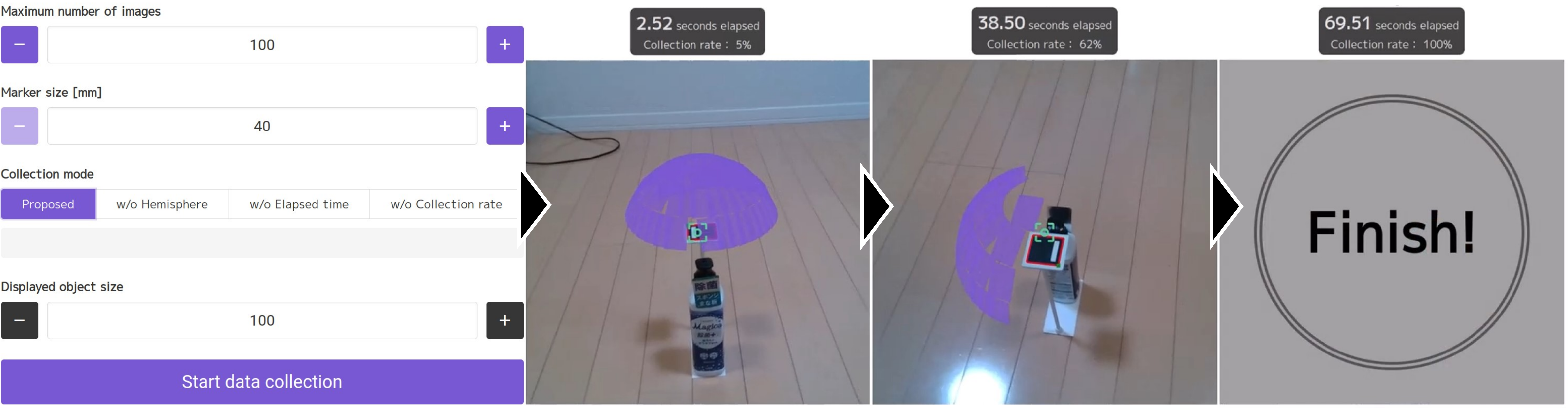}
  \caption{\small{Display transitions of the developed mobile application.}}
  \figlab{transition}
\end{figure*}

To address this issue, Kiyokawa~\etal~proposed a method that uses visual markers as tools to automatically obtain annotations from images~\cite{KiyokawaRAL2019,KiyokawaAR2019}. 
They attached markers to each object and placed the marker-attached objects in different poses in the real world. 
Because a visual marker is easily detected, this method automatically annotates the given positional relationship between an object and the marker. The marker was used for object detection during automatic annotation. However, if an image showing the marker is used for training, normally, the neural network learns the marker as a feature of the object. By hiding the marker with a noise mask, we succeeded in reducing this erroneous learning.

However, even with this method, it is necessary to quickly collect images captured from various directions. Using online visual feedback, we solve this problem by allowing various people to perform this task.
Visual-marker-based automatic annotation allows for consistent annotation even in the presence of occlusions.
In addition to the annotation process, Kiyokawa~\etal~developed a system that automatically changes object poses~\cite{Kiyokawa2021}. Using a hand-eye robot and a rotating stage, the camera fixed on the arm tip can capture objects in various poses by moving the camera and object.

This study endeavors to enhance the overall capacity for collecting large image datasets by improving the level of motivation from enjoyment experienced by individual users, rather than merely introducing intensive human resources.
If efficient image dataset collection can be achieved with an application that any user can use, rather than with a special device, as used in the previous method~\cite{Kiyokawa2021}, there will be a paradigm shift in the dataset collection methodology.
It is easy to develop online visual feedback in web applications because it is an automatic annotation by the AR marker that we developed.

This study aims to develop a method for compactly collecting multi-view object image datasets, which is a web application that allows users to easily capture free-viewpoint images in the real world. 
Users can collect images of the target object by moving a camera device around the target object, which will be automatically annotated later.
\figref{transition} shows an example of the application display showing the collection rate that increases to 100\% (defined as the target number that the user specifies), uncollected viewpoints with a hemisphere-shaped object, and elapsed time from the collection started.

To encourage users to work on this human-in-the-loop dataset collection, enhancing the collection performance and quality of operation is crucial.
Quality of operation in this paper refers to the standard of individual operator's perception that encompasses the emotional, social, and physical aspects of collection operations, including Workload, Enjoyment, and Time pressure.
We implemented the three simple yet effective application features to show dataset collection statuses.
The following items are the proposed application features and \tabref{pros-cons} organizes expectations on collection performance and quality of operation benefitted from the proposals.
\begin{enumerate}
  \item[Hm] \textbf{Hemisphere:} Representing uncollected areas as a hemisphere shape allows users to find where to move the camera, resulting in the collection of unbiased image datasets in an enjoyable manner.
  \item[CR] \textbf{Collection rate:} Displaying the collection rate gives users feedback on the collection's success, leading to motivation from enjoyment of the task.
  \item[ET] \textbf{Elapsed time:} Displaying the elapsed time encourages users to efficiently collect datasets under time pressure even though this may put a little mental strain on the user. Furthermore, the proposed web application aggregated and displayed a ranking of the final elapsed times to gamify the collection work.
  \item[ALL] \textbf{All-functioned}: mode includes the above three features.
\end{enumerate}

For the DL-based object detector dataset, it is necessary to acquire images from various viewpoints to increase the variation in the datasets.
By displaying both types of positions (already captured and must be captured) in the shape of a hemisphere and gamifying the process, where users quickly erase all the surfaces of this hemisphere, the quality of the dataset can be simultaneously improved.
Furthermore, by providing a collection rate during the work and a collection time ranking feature at the end of the application execution, users may perceive the application as a game in which they aim to quickly erase the hemisphere.
\begin{table}[tb]
    \centering
        \small
        \caption{\small{Expected impact of application functions.}}
        \tablab{pros-cons}
        {\tabcolsep = 0.6mm
        \begin{tabular}{p{3mm}p{4mm}p{4mm}p{4mm}p{4mm}p{4mm}} \toprule
            & \multicolumn{2}{c}{Collection performance} & \multicolumn{3}{c}{Quality of operation} \\ \midrule
            & \multicolumn{1}{c}{Efficiency} & \multicolumn{1}{c}{Variation} & \multicolumn{1}{c}{Workload} & \multicolumn{1}{c}{Enjoyment} & \multicolumn{1}{c}{Time pressure} \\ \midrule
            Hm & \multicolumn{1}{c}{\begin{tabular}{c}-\end{tabular}} & \multicolumn{1}{c}{\begin{tabular}{c}\cmark\end{tabular}} & \multicolumn{1}{c}{\begin{tabular}{c}-\end{tabular}} & \multicolumn{1}{c}{\begin{tabular}{c}\cmark\end{tabular}} & \multicolumn{1}{c}{\begin{tabular}{c}-\end{tabular}} \rule[-1mm]{0mm}{4mm} \\
            CR & \multicolumn{1}{c}{\begin{tabular}{c}-\end{tabular}} & \multicolumn{1}{c}{\begin{tabular}{c}-\end{tabular}} & \multicolumn{1}{c}{\begin{tabular}{c}-\end{tabular}} & \multicolumn{1}{c}{\begin{tabular}{c}\cmark\end{tabular}} & \multicolumn{1}{c}{\begin{tabular}{c}-\end{tabular}} \rule[-1mm]{0mm}{4mm} \\ 
            ET & \multicolumn{1}{c}{\begin{tabular}{c}\cmark\end{tabular}} & \multicolumn{1}{c}{\begin{tabular}{c}-\end{tabular}} & \multicolumn{1}{c}{\begin{tabular}{c}\xmark\end{tabular}} & \multicolumn{1}{c}{\begin{tabular}{c}\cmark\end{tabular}} & \multicolumn{1}{c}{\begin{tabular}{c}\cmark\end{tabular}} \rule[-1mm]{0mm}{4mm} \\ \bottomrule
        \end{tabular}
        }
\end{table}

This study examines the impact of the proposed three functions on user's mental workload, usability, motivation from enjoyment, efficiency, and dataset quality. This analysis encompasses the potential positive and negative consequences, as evidenced by the table.
In our experiments, to measure the effect of each function (\eg~Hm, CR, and ET), we conducted an ablation study by comparing it with four methods. One of them is where all three functions were included (ALL) and the other three methods are where each function was removed.

First, seven participants answered the \textit{NASA Task Load Index} (NASA-TLX)~\cite{Hart1988} questionnaire on mental workload, the \textit{System Usability Scale} (SUS)~\cite{Brooke1996} questionnaire on usability, a custom questionnaire on personal information, prior knowledge, prior experiences, and enjoyable features immediately after using the application.
Second, to evaluate the efficiency and possibility of collecting an unbiased dataset, we measured the time taken for the collection trials and calculated the variation of collected datasets to evaluate the efficiency and bias on the dataset.
This study contributes to the literature in two ways:
\begin{enumerate}
  \item To scale up the applicability of automatically annotating object images, we presented, developed, and evaluated a mobile application for collecting multi-view image datasets.
  Using the developed system, this study clarified the feasibility of annotation and object detection (\textbf{Section~6}).
  \item Thorough evaluations clarified the possibility of incorporating the proposed functions (Hm, CR, and ET) can lead to dataset collection with less mental workload, less time pressure, and high dataset variation in an enjoyable manner (\textbf{Sections~5 and 7}).
\end{enumerate}

\section{Related Works}
This section describes related studies on online visual feedback techniques for effective training dataset collection and the benefits of cloud-based dataset collection.

\subsection{Online Visual Feedback in Collecting Dataset}
In the field of computer vision, to annotate proper labels for images on the web toward more accurate image search, Ahn~\etal~\cite{Ahn2004} introduced an interactive system: a game that is fun and can be used to provide meaningful labels for images on the web.
Deng~\etal~\cite{Deng2013} presented a game known as \textit{bubbles} that reveals discriminative image features used by humans. The player's goal is to identify the category of heavily blurred images. They further proposed an image recognition algorithm, \textit{Bubble Bank}, which uses human-selected bubbles to improve the recognition performance. 

Similar to our purpose, Kavasidis~\etal~\cite{Kavasidis2013} presented a flash game that aims to easily generate the ground truth for testing fish detectors.
They developed an online game known as \textit{Flash the Fish} in which the user is shown videos from underwater environments and has to take photos of the fish by clicking on them.
This method relies on manually clicking on fish in images.
Comparing the results obtained to a hand-drawn ground truth confirmed that a game that implicitly controls user quality can easily create reliable images and video annotations.
However, this game is limited to fish detection, and the game creator must invest significant efforts in designing game scenarios for each domain-specific detection model.

In another study addressing dataset creation for object detection, Fusiello~\etal~\cite{Fusiello2012} proposed an approach to detect and label objects within images. They described a two-player web-based guessing game called \textit{Ask’nSeek}, which supports these tasks in a fun and interactive manner.
Ask’nSeek requires users to guess the location of a hidden region within an image using semantic and topological clues; thus, this is a semi-automatic annotation method unlike ours.
The information collected from game logs was combined with the results of image content analysis algorithms and used to feed a machine-learning algorithm that outputs the outline of the most relevant regions within the image and their names.

They clarified the feasibility of games used for image annotations. However, as aforementioned, these methods~\cite{Fusiello2012,Kavasidis2013} are incapable of automatically annotating real-world images.

\subsection{Cloud-based Data Sharing}
Several researchers have studied the potential benefits of \textit{cloud robotics} technologies such as big data, cloud computing, collective robot learning, and human computation, as discussed and surveyed in~\cite{Hu2012,Kehoe2015,Chen2018}.

In the field of robotic manipulation, grasping~\cite{Li2018} and vision systems~\cite{Klank2009,Bistry2010} that utilize cloud computing and Internet services have been proposed thus far.
Previous studies demonstrated the feasibility of practical cloud-based robotic applications.
One remaining important issue is the construction of an effective dataset collection method for training systems with a particular affinity for the cloud robotics framework.
Without a dataset collection method that can be easily and quickly uploaded to the Internet, we cannot assume that a large number of databases are available on the Internet; thus, responding promptly when various new objects appear is highly difficult.
Among these dataset collection problems, this study focuses on object image dataset collection, assuming a cloud-based robot vision system.

\section{Human-in-the-loop Dataset Collection}
\subsection{System Implementation}
\begin{figure}[tb]
    \centering
    \includegraphics[keepaspectratio, width=\linewidth]{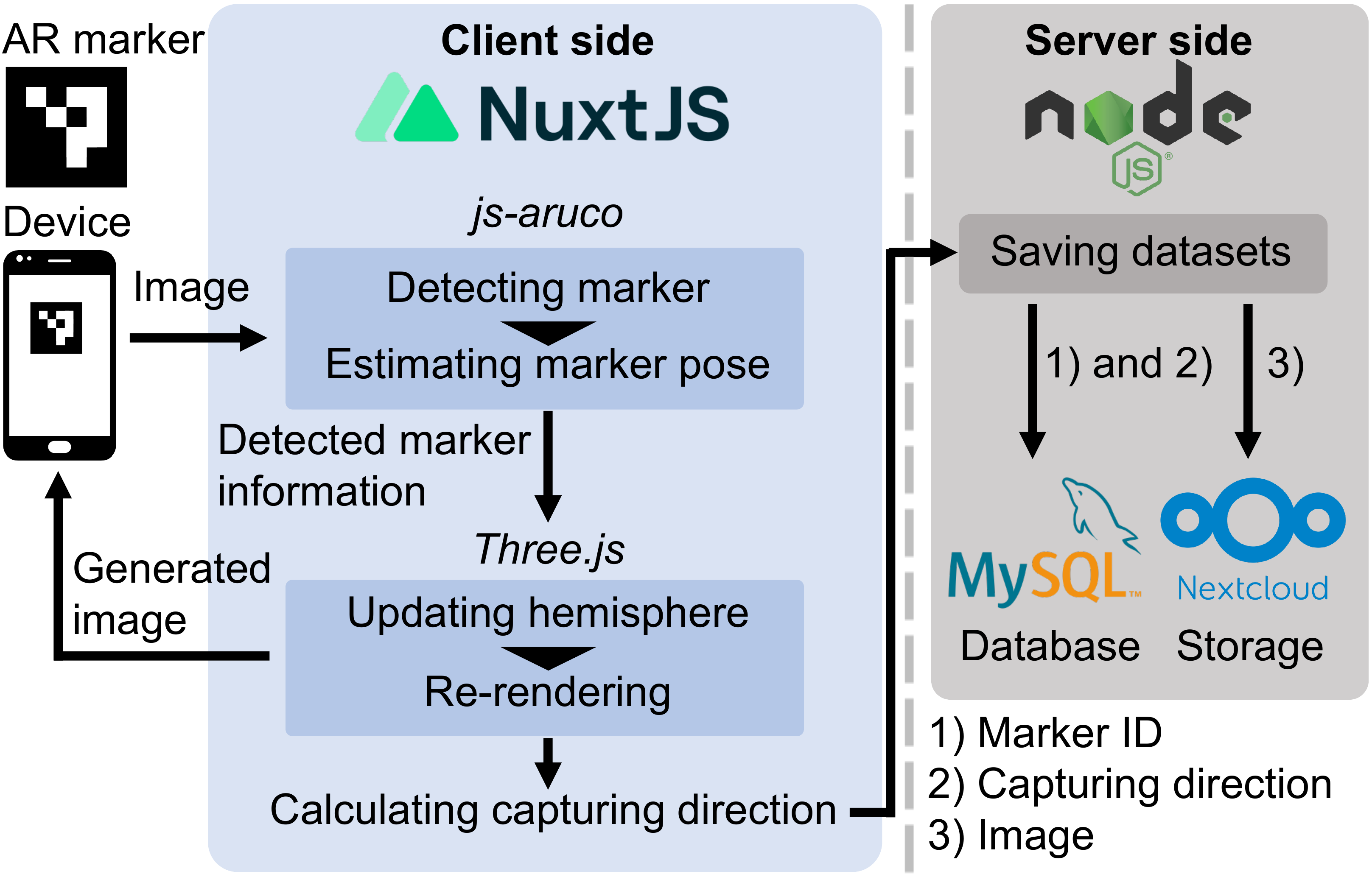}
    \caption{\small{System configuration for mobile application.}}
    \figlab{system}
\end{figure}
\begin{figure}[tb]
    \centering
    \begin{minipage}[tb]{0.51\linewidth}
        \centering
        \includegraphics[keepaspectratio, width=\linewidth]{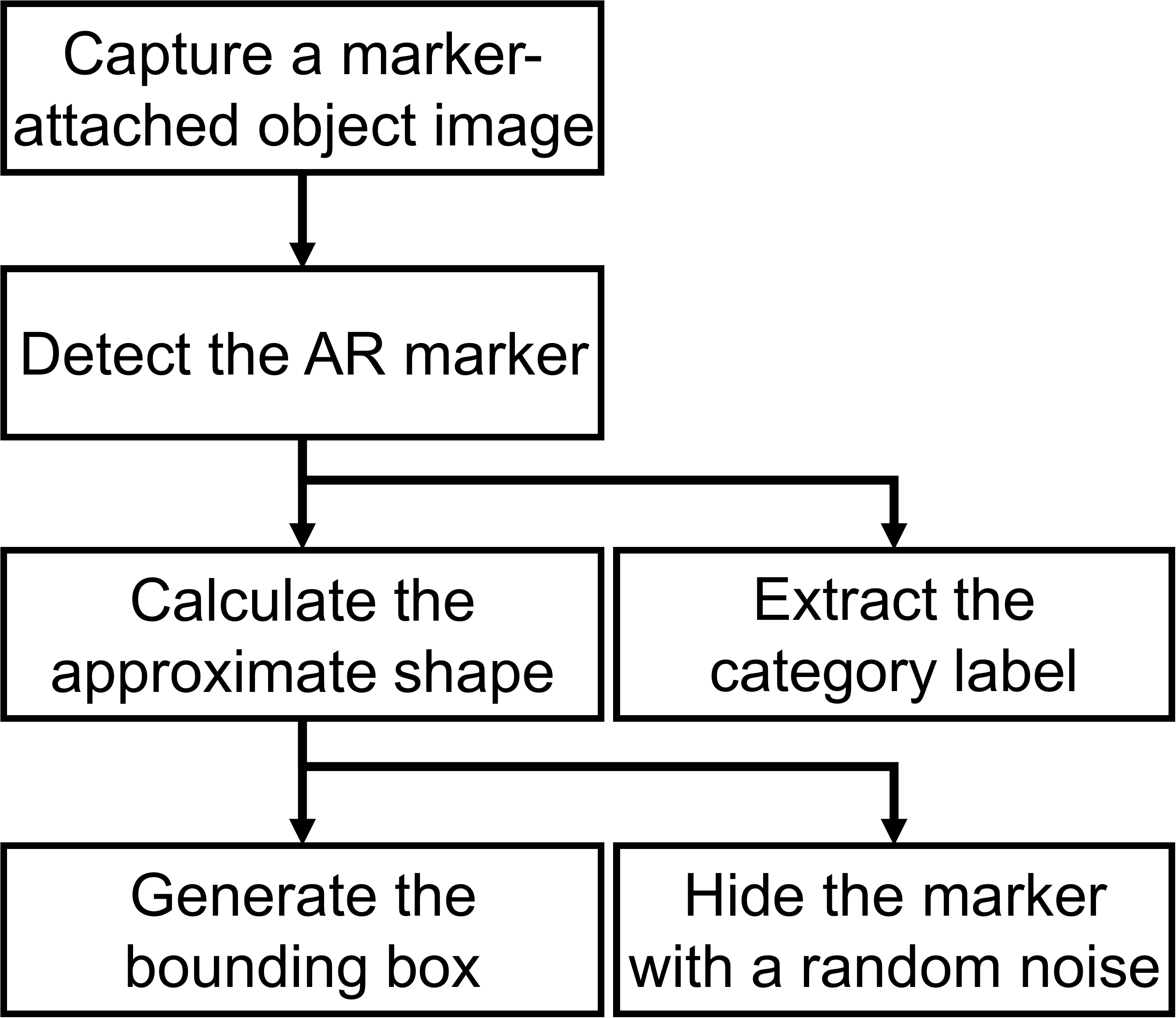}
        \subcaption{\small{Annotation flow}}
    \end{minipage}
    \begin{minipage}[tb]{0.47\linewidth}
        \centering
        \includegraphics[keepaspectratio, width=\linewidth]{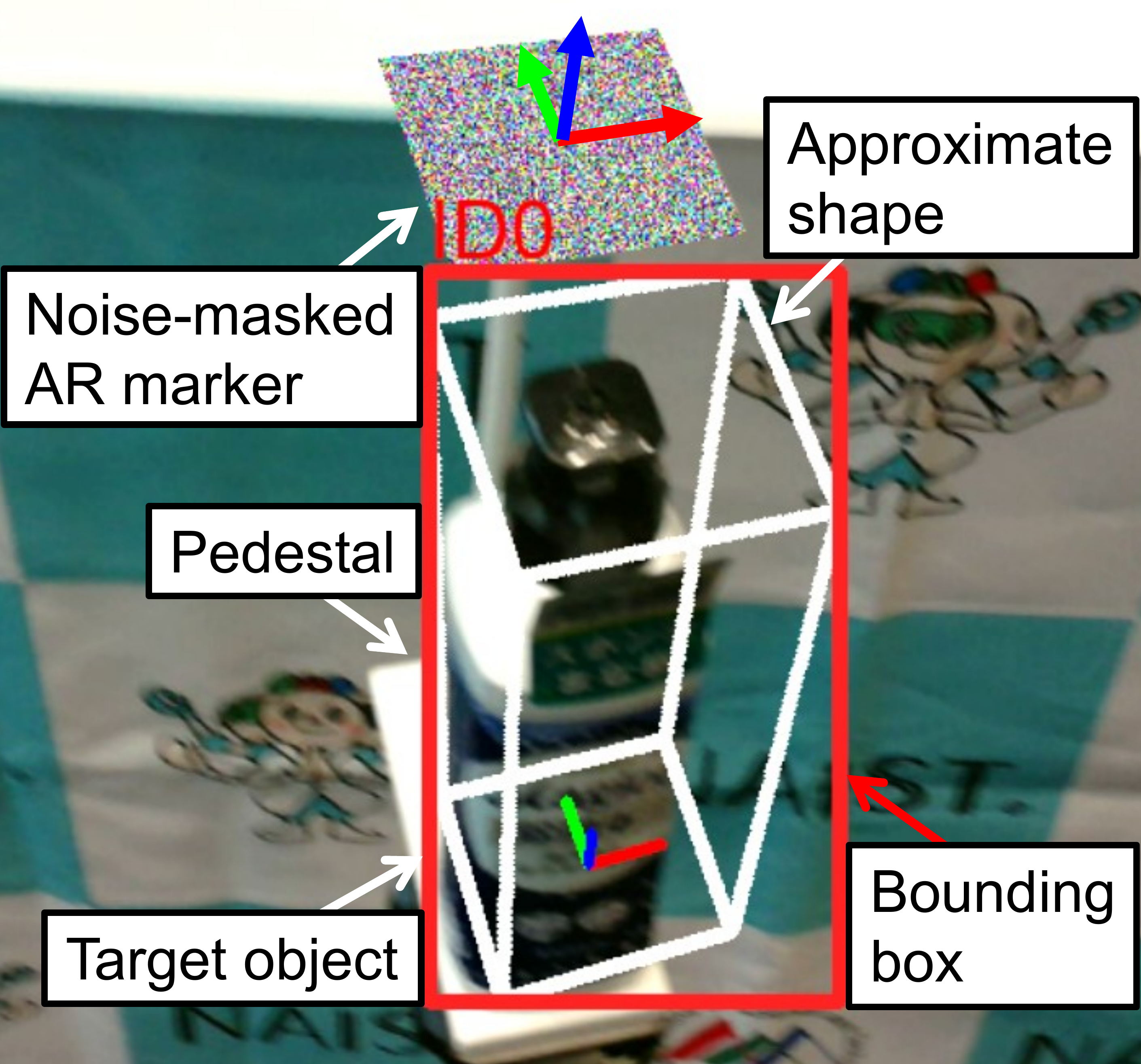}
        \subcaption{\small{Superimposed annotation}}
    \end{minipage}
    \caption{\small{Automatic 2D bounding box annotation process.}}
    \figlab{annotation}
\end{figure}
\begin{figure}[tb]
  \centering
  \includegraphics[width=\linewidth]{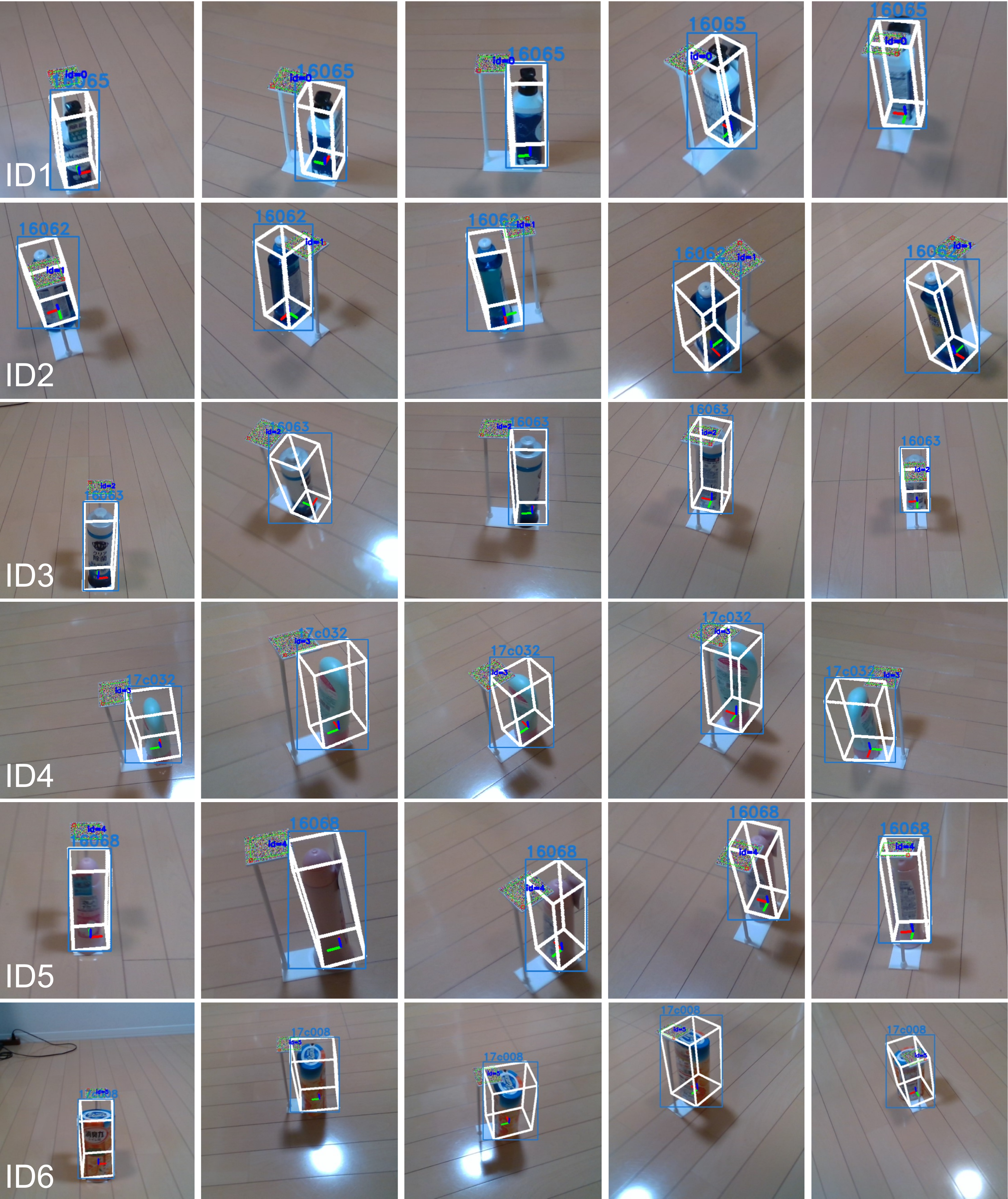}
  \caption{\small{Visualization of collected datasets.}}
  \figlab{collected}
\end{figure}

\figref{system} shows the system configuration of the mobile application. 
We implemented a web application consisting of a client-side component for collecting and annotating images and visualizing the collection status and a server-side component for storing the resulting datasets.
In general, web applications are device-independent, including mobile devices such as Windows tablets and Android smartphones.
These mobile devices often come with an additional camera, and the web application installed on the device allows for easy capture of multi-view images from free viewpoints by moving the device’s camera. The computational resources of the device are used to perform image acquisition, marker detection, marker pose estimation, and various drawings on the display.

To automate the 2D bounding box annotation for object images, we employed an AR marker-based annotation method~\cite{KiyokawaRAL2019}.
\figref{annotation}\textbf{(a)} shows the process flow.
Before starting the application, the user places the marker-attached jig by fixing the target object at the location where it can be captured by the camera from around the object. 
The participants used a pre-calibrated camera. The calibration was performed using an NxM checkerboard and OpenCV calibrateCamera method.
Because a known relative pose between the object and marker on the jig, and marker identification numbers (IDs) associated with the object category are assumed, an automatic annotation process can be performed during the execution of the web application.
\figref{annotation}\textbf{(b)} shows a visualization of the collected dataset, whereas \figref{collected} shows more examples of different objects captured from different viewpoints.

We use \textit{Nuxt.js}\footnote{https://nuxtjs.org/} with several JavaScript libraries to fulfill various functions, including augmented reality (AR) library, \textit{js-aruco}\footnote{https://github.com/jcmellado/js-aruco}, and 3D library, \textit{Three.js}\footnote{https://threejs.org/}. 
The js-aruco detects AR markers in captured images and estimates the poses of the markers.
The three.js is a user-friendly library that can render 3D content on a web page, which in turn provides support for updating the appearance of the hemisphere in captured images by updating the pose, removing the rectangle on the hemisphere just captured, and applying it to render again.
Finally, the rendered image was shown on the display of the mobile device.
During one loop on the client side, the server saves the image dataset on cloud storage structured by the \textit{MySQL}\footnote{https://www.mysql.com/} database and \textit{NextCloud}\footnote{https://nextcloud.com/} online storage services.

\figref{transition} shows an example of display transitions.
On the initial display, users set four parameters: the target number of images, the size of the visual marker attached to the object, the collection mode to be executed, and the size of the hemispherical object to be displayed.
Once these parameters are set, users can begin the collection process. As soon as the selected mode starts, the user can view the captured image by displaying the proposed status indicators, including the collection progress as a hemisphere (Hm), collection rate (CR), and elapsed time (ET). Once the user-specified number of images has been collected, image capture automatically ends.

\subsection{Online Visual Feedback by Visualizing Status}
The display at the starting point shows the descriptions of the selected mode and the parameters required to activate the mode.
When clicking the button labeled as \textit{start data collection} shown at the bottom of the leftmost picture of~\figref{transition}, count down until image capture begins.
After counting from five seconds to zero, the image capture and marker detection began simultaneously.
During image capture, mode-specific visualizations, which include the ET, CR, and Hm showing the collection progress, appear on the display.
When the collection rate reached 100\%, image capture and marker detection ended, and the message \textit{Finish!} was displayed on the display.
After finishing the image capture, users can observe the collection time ranking in the top window of the web application.

The following sections provide more details on the proposed three application features (\tabref{pros-cons}) of displaying the elapsed time, showing the collection rate, and representing the uncollected areas as a hemisphere.

\subsubsection{Drawing a Collection Progress as Hemisphere}
To implicitly show collection progress to the user during the collection trial, the application window shows a hemisphere, that has missing parts, indicating uncaptured areas.
The hemisphere is divided into a certain number of rectangular shapes (the number can be set in the form of the \textit{displayed object size} shown in the leftmost image of~\figref{transition}), and the divided rectangular objects are displayed first.
To arrange the rectangles such that they do not overlap with each other, the placement of the rectangular objects forms a spiral on the hemisphere surface, as shown in~\figref{transition}.
The rectangles on the hemisphere were placed as described in~\cite{Saff1997}.

Using \textit{setFromSphericalCoords} in \textit{Vector3} class\footnote{https://threejs.org/docs/\#api/en/math/Vector3} of the three.js, we can easily calculate an object position $\bm{v}$ in the Cartesian coordinate system as follows.
\begin{equation}
\bm{v} = \left(r\sin\phi\sin\theta, r\cos\phi, r\sin\phi\cos\phi \right),
\end{equation}
where $r$ denotes the hemispherical radius.
To set the planar normal of the rectangular objects to be parallel to $\bm{v}$, we use a function \textit{lookAt} of \textit{Object3D} class\footnote{https://threejs.org/docs/\#api/en/core/Object3D} of the three.js.

Once one of the rectangular objects exists in the center of the captured image, the corresponding displayed rectangle disappears indicating the capturing is finished.
The hemisphere hides as little of the object's appearance as possible. A semi-transparent hemispherical object is superimposed on the image displayed in rendering. These semi-transparent objects allow the user to easily observe the actual target object's pose.
Users can monitor the progress of the collection not only through the collection rate but also by observing the updating hemisphere, which provides information about the remaining uncaptured areas and helps the user move the camera effectively.
With this feature, the viewpoints (\ie~the object pose observed from the camera) do not overlap with existing datasets, leading to unbiased datasets.

Displaying the hemisphere provides users feedback during the collection process.
This leads to a feeling of collection, which is similar to a game. 
Users can enjoy collecting images while considering methods to remove the hemisphere as quickly as possible.
Such game-like functions motivate users to collect datasets quickly.

\subsubsection{Displaying Numerical Collection Percentage}
\figref{transition} shows the appearance of the displayed collection rate during the execution.
The calculated collection rate is shown at the top of the display above the image capture window.
Displaying the collection rate also leads to a feeling of collection, similar to a game. 
Users can enjoy collecting images while considering methods to increase their collection rate. 

The collection rate was calculated based on six-dimensional (6D) object pose (with respect to the camera viewpoint) data included in the previously collected datasets.
In reality, once the rectangle shown on the hemisphere surface is located at the center of the image captured by the camera, the target rectangle disappears and the obtained image dataset is saved.
Therefore, the collection rate increases by 1\% in the case of 100 image collection.
This method can reduce the duplication of datasets from the camera viewpoint.
By enabling users to avoid this dataset duplication on their own, this feature creates comfort and playability.

The drawing of the hemisphere shows this progress in more detail and intuitively.
In contrast to the hemisphere drawing, this collection rate can be a quantitative measure of how much has now been collected against the final target value (100\%).

\subsubsection{Providing Elapsed Time and Ranking}
The elapsed time displayed during mode execution encourages users to collect the dataset more quickly under time pressure, even though this may put a little mental strain on the user.
Furthermore, the proposed web application aggregated and displayed a ranking of the final elapsed times to motivate users to enjoyable and efficient collection like a game.
Shorter collection times are considered higher in the rankings.
In the ranking table, we can observe the rank of the previous trial in all trials, mode name, performance (capture time [s] / number of images), capture time [s], and a number of images.

\section{Evaluating User Experiences}
\subsection{Experimental Protocol}
Our user experiments evaluate the mental workload, usability, motivation from enjoyment, efficiency (collection time), and dataset quality (variation) of the dataset collection trials using the proposed application.
One application (ALL) is equipped with three functions (\ie~Hm, CR, and ET), which were implemented in a single application mode.
We developed other three application modes, each lacking one of the three functions: a mode without Hm, a mode without CR, and a mode without ET.
The typical procedure for all implemented modes to collect datasets for a target object is as follows:
\begin{enumerate}
 \setlength{\parskip}{0cm} 
 \setlength{\itemsep}{0cm} 
 \item Attach a visual marker at a location near the target object where the object is easily detectable
 \item Launch the web application on a small handheld laptop (Microsoft, Surface PRO)
 \item Choose one mode of the four modes (We randomly generated the order for each participant beforehand)
 \item Start the mode that captures multi-view images while checking the visualizations on the display
 \item After that the mode ends (if the collection rate reaches 100\%), and the participant checks the ranking together with other collection results
 \item The participant takes questionnaires including NASA-TLX, SUS, and custom questionnaires
 \item The series of processes from 3) to 6) repeats each of the four modes three times
\end{enumerate}

We surveyed seven male adults from the Nara Institute of Science and Technology, Japan.
Participants completed the questionnaire, as shown in~\tabref{questionnaire}.
We were limited to experiments with participants collected from the same university because of the countermeasures against COVID-19. However, as much as possible, we conducted experiments by collecting subjects from different backgrounds.
As the responses to the questions are shown in~\tabref{bg}, we recruited participants with different specialties.
\tabref{exp-seq} shows the experiment sequence for each participant.

Once 100 images are collected after starting the mode, the collection rate reaches 100\%, and the executed mode ends.
As shown in~\figref{transition}, the size of the visual marker attached to the object and the size of the hemispherical object to be displayed were set to 40mm (length of one side) and 100mm (radius), respectively.
\renewcommand{\arraystretch}{1.2}
\begin{table*}[tb]
\begin{threeparttable}[tb]
    \small
    \centering
        \caption{\small{Custom-made questionnaires prepared separately from NASA-TLX and SUS.}} \tablab{questionnaire}
        {\tabcolsep = 1.0mm
        \begin{tabular}{p{95mm}p{21mm}p{54mm}} \toprule
            \multicolumn{1}{c}{Question} & \multicolumn{1}{c}{Type} & \multicolumn{1}{c}{Option} \\ \midrule
            Gender & Multiple-choice & Male or Female \\
            Age & Open-ended & Positive integer \\
            Have you ever experienced any AR technologies? & Closed & No or Yes \\
            How much do you know about training dataset collection? & Multiple-choice & Nothing, A little, Well, or Your major \\
            Did you enjoy it like a game? Which functions did you find enjoyable? & Multiple-choice & Not enjoy it, Hm, CR, ET, or Ranking \\
            Did you try hard to collect the dataset? Which functions motivated you? & Multiple-choice & Don't think so, Hm, CR, ET, or Ranking \\ \bottomrule
        \end{tabular}
        }
\end{threeparttable}
\end{table*}
\renewcommand{\arraystretch}{1.0}
\begin{table}[tb]
\begin{threeparttable}[tb]
    \small
    \centering
        \caption{\small{Responses about participant's background.}} \tablab{bg}
        {\tabcolsep = 2.4mm
        \begin{tabular}{p{14mm}p{11mm}p{11mm}p{11mm}p{11mm}p{11mm}p{11mm}p{11mm}} \toprule
            & \multicolumn{7}{c}{Participant ID} \\ \cmidrule(r){2-8}
            \multicolumn{1}{c}{Question ID} & \multicolumn{1}{c}{1} & \multicolumn{1}{c}{2} & \multicolumn{1}{c}{3} & \multicolumn{1}{c}{4} & \multicolumn{1}{c}{5} & \multicolumn{1}{c}{6} & \multicolumn{1}{c}{7} \\ \midrule
            1$^{\rm \dag{a}}$ & \multicolumn{1}{c}{No} & \multicolumn{1}{c}{No} & \multicolumn{1}{c}{Yes} & \multicolumn{1}{c}{Yes} & \multicolumn{1}{c}{Yes} & \multicolumn{1}{c}{Yes} & \multicolumn{1}{c}{Yes} \\
            2$^{\rm \dag{b}}$ & \multicolumn{1}{r}{4} & \multicolumn{1}{r}{2} & \multicolumn{1}{r}{2} & \multicolumn{1}{r}{3} & \multicolumn{1}{r}{2} & \multicolumn{1}{r}{4} & \multicolumn{1}{r}{4} \\ \bottomrule
        \end{tabular}
        }
        \vspace{1mm}
        \begin{tablenotes}
          \item[$\rm \dag{a}$]\footnotesize{Have you ever experienced any AR technologies?}
          \item[$\rm \dag{b}$]\footnotesize{How much do you know about training dataset collection? The response numbers 1, 2, 3, and 4 correspond to Nothing, Know a little, Know well, and Your major, respectively.}
        \end{tablenotes}
\end{threeparttable}
\end{table}
\begin{table}[tb]
    \centering
        \small
        \caption{\small{Experiment sequence for each participant.}}
        \tablab{exp-seq}
        {\tabcolsep = 0.6mm
        \begin{tabular}{p{15mm}p{18mm}p{18mm}p{18mm}p{18mm}} \toprule
            & \multicolumn{4}{c}{Experiment sequence} \\ \cmidrule(r){2-5}
            \multicolumn{1}{c}{Participant ID} & \multicolumn{1}{c}{1} & \multicolumn{1}{c}{2} & \multicolumn{1}{c}{3} & \multicolumn{1}{c}{4} \\ \midrule
            \multicolumn{1}{c}{1} & \multicolumn{1}{c}{w/o ET$\times$3} & \multicolumn{1}{c}{w/o CR$\times$3} & \multicolumn{1}{c}{ALL$\times$3} & \multicolumn{1}{c}{w/o Hm$\times$3} \rule[-1mm]{0mm}{4mm} \\
            \multicolumn{1}{c}{2} & \multicolumn{1}{c}{w/o Hm$\times$3} & \multicolumn{1}{c}{w/o CR$\times$3} & \multicolumn{1}{c}{ALL$\times$3} & \multicolumn{1}{c}{w/o ET$\times$3} \rule[-1mm]{0mm}{4mm} \\ 
            \multicolumn{1}{c}{3} & \multicolumn{1}{c}{w/o Hm$\times$3} & \multicolumn{1}{c}{w/o ET$\times$3} & \multicolumn{1}{c}{w/o CR$\times$3} & \multicolumn{1}{c}{ALL$\times$3} \rule[-1mm]{0mm}{4mm} \\ 
            \multicolumn{1}{c}{4} & \multicolumn{1}{c}{w/o CR$\times$3} & \multicolumn{1}{c}{w/o ET$\times$3} & \multicolumn{1}{c}{w/o Hm$\times$3} & \multicolumn{1}{c}{ALL$\times$3} \rule[-1mm]{0mm}{4mm} \\ 
            \multicolumn{1}{c}{5} & \multicolumn{1}{c}{ALL$\times$3} & \multicolumn{1}{c}{w/o Hm$\times$3} & \multicolumn{1}{c}{w/o ET$\times$3} & \multicolumn{1}{c}{w/o CR$\times$3} \rule[-1mm]{0mm}{4mm} \\ 
            \multicolumn{1}{c}{6} & \multicolumn{1}{c}{w/o ET$\times$3} & \multicolumn{1}{c}{ALL$\times$3} & \multicolumn{1}{c}{w/o CR$\times$3} & \multicolumn{1}{c}{w/o Hm$\times$3} \rule[-1mm]{0mm}{4mm} \\ 
            \multicolumn{1}{c}{7} & \multicolumn{1}{c}{w/o CR$\times$3} & \multicolumn{1}{c}{ALL$\times$3} & \multicolumn{1}{c}{w/o Hm$\times$3} & \multicolumn{1}{c}{w/o ET$\times$3} \rule[-1mm]{0mm}{4mm} \\ \bottomrule
        \end{tabular}
        }
\end{table}

\subsection{Evaluation Metrics}
\textbf{Section~5} discusses the results in several aspects of these results.
First, we analyzed the results of the questionnaires to evaluate mental workload (\textbf{Section~5.1}) and usability (\textbf{Section~5.2}) from a subjective perspective.
Referring to~\cite{Shirakura2021}, we asked the participants several questions regarding their usage during the trials to compare the effects of each feature implemented in each mode.
To evaluate mental workload and usability, the seven participants took the NASA-TLX and SUS respectively.
To clarify the contributions of the implemented functions to the motivation from enjoyment of users, the participants answered the custom questionnaires listed in~\tabref{questionnaire} (\textbf{Section~5.3}).
Second, we evaluated the efficiency of the proposed web application in terms of collection time (\textbf{Section~5.4}) together with the variation of the collected dataset (\textbf{Section~5.5}).

\section{Results}
\subsection{Mental Workload}
\begin{figure}[tb]
  \centering
  \includegraphics[width=\linewidth]{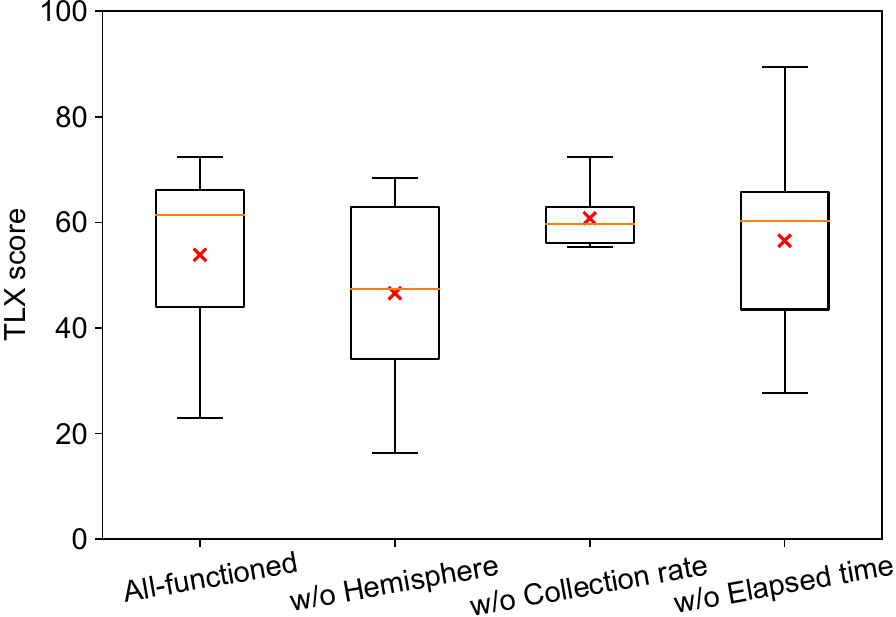}
  \caption{\small{TLX score.}}
  \figlab{tlx}
\end{figure}
\begin{figure}[tb]
  \centering
  \includegraphics[width=\linewidth]{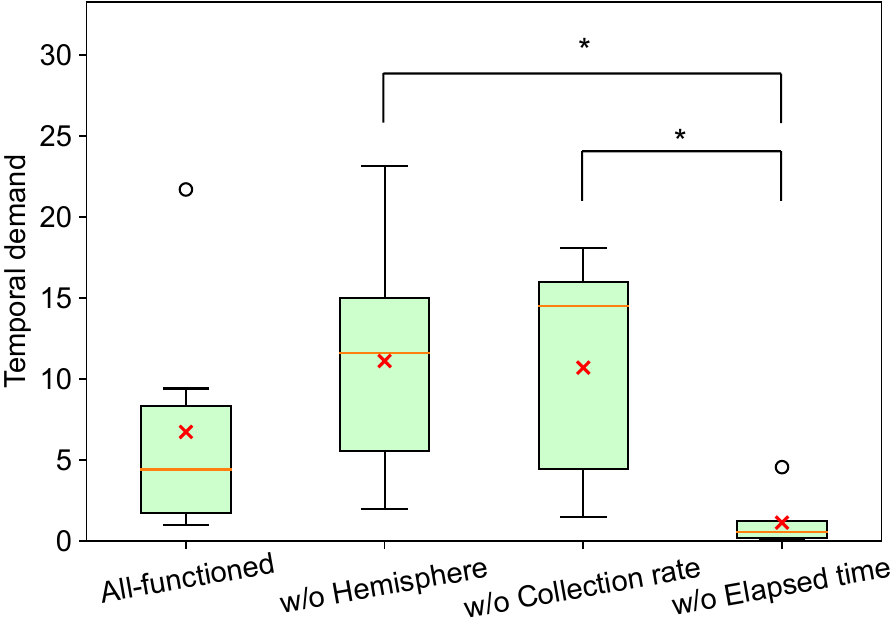}
  \caption{\small{``Temporal demand'' (a NASA TLX subscale).}}
  \figlab{td}
\end{figure}
\begin{figure}[tb]
  \centering
  \includegraphics[width=\linewidth]{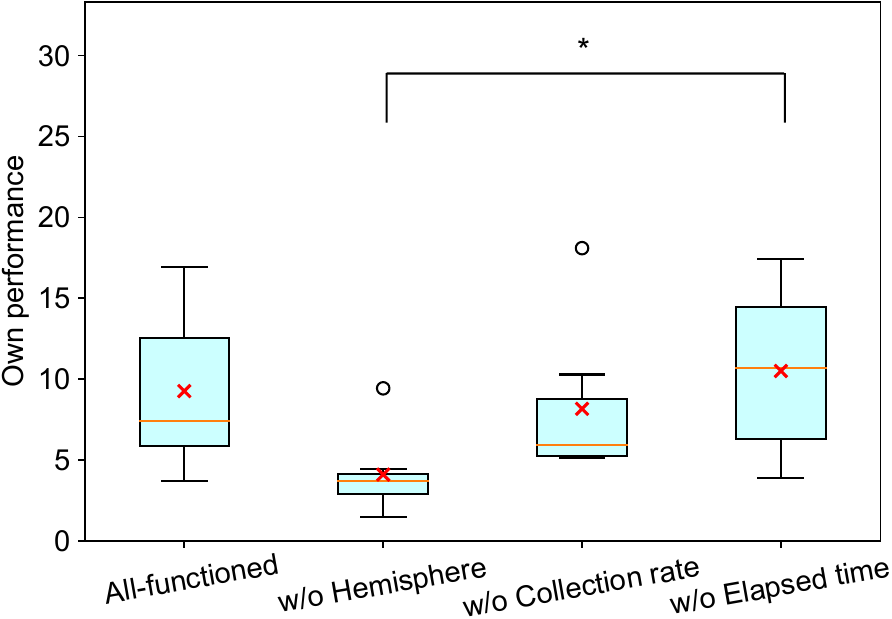}
  \caption{\small{``Performance'' (a NASA TLX subscale).}}
  \figlab{op}
\end{figure}
\begin{figure}[tb]
  \centering
  \includegraphics[width=\linewidth]{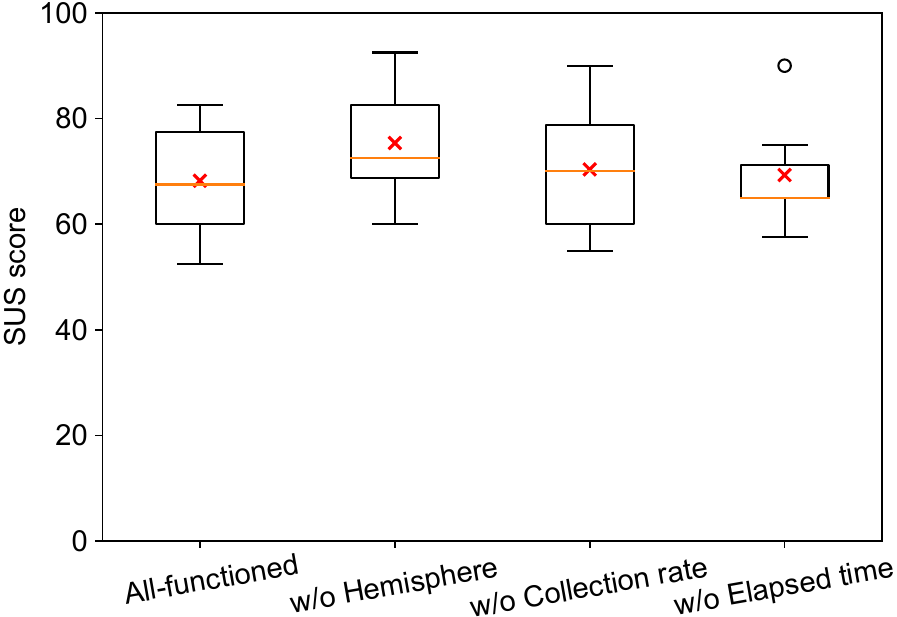}
  \caption{\small{SUS score.}}
  \figlab{sus}
\end{figure}
\begin{figure}[tb]
  \centering
  \includegraphics[width=\linewidth]{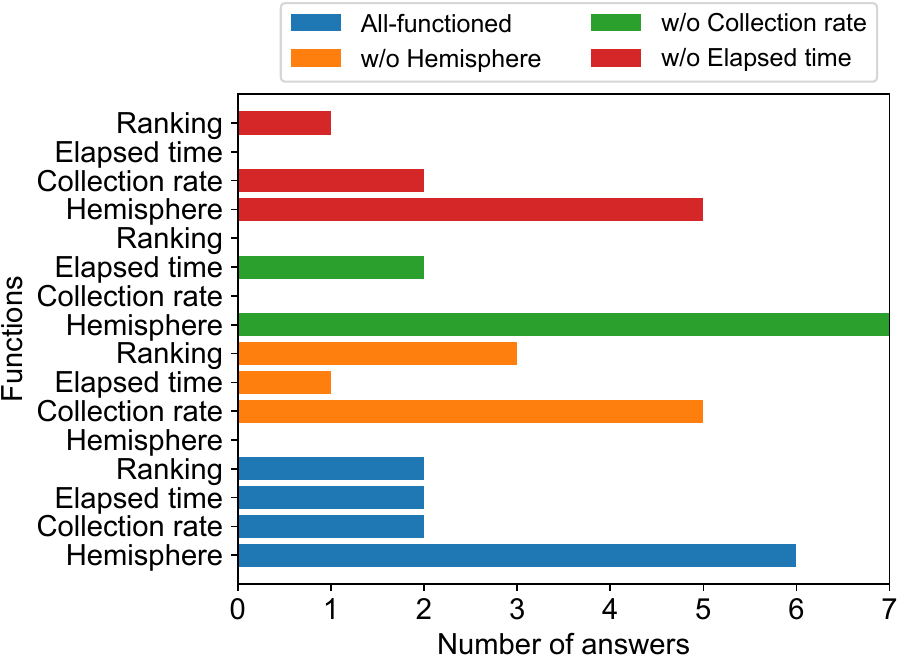}
  \caption{\small{Count of responses to custom questions ``Did you enjoy it like a game?" and ``Which features did you find enjoyable?''.}}
  \figlab{count_enjoy}
\end{figure}
\begin{figure}[tb]
  \centering
  \includegraphics[width=\linewidth]{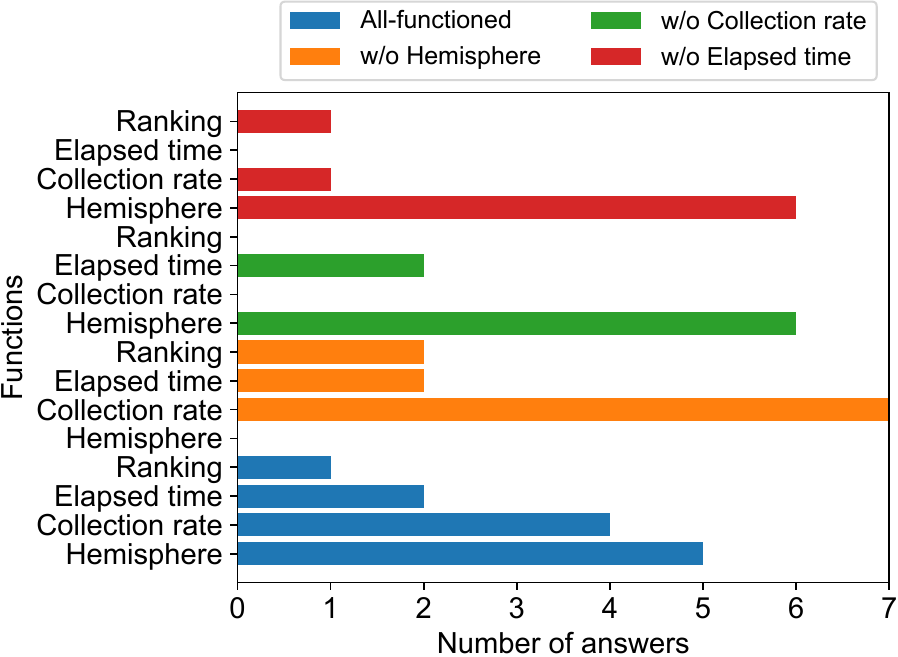}
  \caption{\small{Count of responses to custom questions ``Did you try hard to collect the dataset?" and ``Which features motivated you?''.}}
  \figlab{count_motivate}
\end{figure}
\figref{tlx} shows the results of the NASA-TLX questionnaire. The TLX score expresses the users' mental workload as a value ranging from 0 to 100, with a larger value indicating a higher workload.
A lower value indicates better performance.
\figref{td} and \figref{op} show scores of two different subscales of NASA-TLX, which include \textit{Temporal demand} and \textit{Performance}.

The red-colored ``x'' mark indicates the average value, and the upper and lower limits of the error bars indicate the maximum and minimum values, respectively.
The top line of the box represents the 75th percentile, the horizontal line represents the median, and the bottom line represents the 25th percentile. 
Samples greater than the 75th percentile $+ 1.5 \times$ the interquartile range (IQR $=$ 75th percentile $-$ 25th percentile) or less than the 25th percentile $- 1.5 \times$ IQR are regarded as outliers (white circles).

Moreover, multiple comparison tests were conducted to compare the questionnaire results.
\tabref{multi_comp} shows the result of multiple comparisons on NASA-TLX subscales.
To conduct multiple comparisons, the \textit{Dunn-Bonferroni test}~\cite{Dunn1964} was used in the statistical software \textit{SPSS}\footnote{https://www.ibm.com/spss}.
The p-values shown in~\tabref{multi_comp} are values corrected by Bonferroni correction with an upper limit of $1.0$.
The asterisked p-values indicate a significant difference between methods 1 and 2 at a significance level of $0.05$, which is also shown in~\figref{td} and \figref{op}

We found several differences in the scores of subscales of NASA-TLX, particularly for \textit{temporal demand} and \textit{performance}.
The method without Hm and the method without CR were rated higher scores of subscale \textit{temporal demand} than the method without ET, as the p-values were $0.010$ and $0.010$, suggesting that in the two methods, users felt that the temporal demand was high.
No significant difference was confirmed between the ALL method and the method without ET, with a p-value of $0.14$. However, all seven subjects reported higher temporal demand scores for the ALL method than those without ET. This result indicates that one of the factors related to mental workload was relatively high in the method with ET.

Although no significant difference could be confirmed between the ALL method and the method without Hm and CR, it was confirmed that five participants felt a higher temporal demand for the methods without Hm and without CR than for the ALL method.
This result shows the possibility of suppressing the temporal demand by combining online visual feedback functions such as the method with Hm and CR in the ALL method.
Both functions served the purpose of displaying the progress of the collection operation.
As Myers~\cite{Myers1985} demonstrated, progress indicators are generally preferred by users, as they provide novices with confidence that a task has been accepted and is progressing successfully, while expert users can obtain sufficient information to predict the approximate completion time of the task.
It is possible that the user's sense of relief, resulting from the progress indicators, also worked to reduce temporal demand during the mobile application-based dataset collection.

The method without Hm showed a significantly lower score on the subscale \textit{performance} than the method without ET, with a p-value of $0.030$.
The subscale \textit{performance} shows the degree of users' self-evaluation of their collection trials.
A higher score indicates a poorer self-evaluation.
Therefore, the results indicate that the self-evaluation of performance tended to be better without Hm, compared with the method without ET. This improvement in self-evaluation was likely due to the explicit visual feedback of the collection outcomes provided by the real-time erasure of the hemisphere.

\subsection{Usability}
\figref{sus} shows the results of the SUS questionnaire.
The SUS score expresses the usability evaluation by users as a value from 0 to 100, with a larger value indicating higher usability.
The format of the graph shown in~\figref{sus} is the same as other figures (\figref{tlx}~-~\figref{op}).
\tabref{multi_comp} also shows the results of multiple comparisons on SUS scores.
There are no significant differences among the methods.

\subsection{Motivation from Enjoyment}
\figref{count_enjoy} and \figref{count_motivate} illustrate the results of the custom-made questionnaires about motivation from enjoyment.
\figref{count_enjoy} shows the count of responses to custom questions ``Did you enjoy it like a game?" and ``Which features did you find enjoyable?''. \figref{count_motivate} shows the count of responses to custom questions ``Did you try hard to collect the dataset?" and ``Which features motivated you?''.
Updating Hm was the most selected because it facilitates enjoyment and is easily motivated. In the two types of questions, the notification of collection rates to users was the second most selected among the four methods.
The other custom-made questions listed in~\tabref{questionnaire} were unremarkable, but the ranking and elapsed time also made the users fun and motivated.

As expected, these results suggest that the display of the updating Hm and CR might be effective in online visual feedback of the dataset collection in terms of motivation from enjoyment.
\begin{table*}[tb]
\begin{threeparttable}[tb]
    \small
    \centering
        \caption{\small{Multiple comparison test.}} \tablab{multi_comp}
        \begin{tabular}{p{13mm}p{13mm}p{18mm}p{13mm}p{13mm}p{13mm}p{13mm}p{13mm}p{13mm}p{13mm}} \toprule
            & & \multicolumn{8}{c}{p-value} \\ \cmidrule(r){3-10}
            \multicolumn{1}{c}{~Method 1~} & \multicolumn{1}{c}{~Method 2~} & \multicolumn{1}{c}{NASA-TLX} &  \multicolumn{1}{c}{~~~MD$^{\rm \dag{a}}$~~~} & \multicolumn{1}{c}{~~~PD$^{\rm \dag{b}}$~~~} & \multicolumn{1}{c}{~~~TD$^{\rm \dag{c}}$~~~} & \multicolumn{1}{c}{~~~OP$^{\rm \dag{d}}$~~~} & \multicolumn{1}{c}{~~~EF$^{\rm \dag{e}}$~~~} & \multicolumn{1}{c}{~~~FR$^{\rm \dag{f}}$~~~} & \multicolumn{1}{c}{~~~~SUS~~~~} \\ \midrule

            \multicolumn{1}{l}{ALL} & \multicolumn{1}{l}{w/o CR} & \multicolumn{1}{r}{1.0} & \multicolumn{1}{r}{1.0} & \multicolumn{1}{r}{0.72} & \multicolumn{1}{r}{1.0} & \multicolumn{1}{r}{1.0} & \multicolumn{1}{r}{1.0} & \multicolumn{1}{r}{1.0} & \multicolumn{1}{r}{1.0} \\

            \multicolumn{1}{l}{ALL} & \multicolumn{1}{l}{w/o ET} & \multicolumn{1}{r}{1.0} & \multicolumn{1}{r}{1.0} & \multicolumn{1}{r}{1.0} & \multicolumn{1}{r}{0.14} & \multicolumn{1}{r}{1.0} & \multicolumn{1}{r}{1.0} & \multicolumn{1}{r}{1.0} & \multicolumn{1}{r}{1.0} \\

            \multicolumn{1}{l}{ALL} & \multicolumn{1}{l}{w/o Hm} & \multicolumn{1}{r}{1.0} & \multicolumn{1}{r}{1.0} & \multicolumn{1}{r}{1.0} & \multicolumn{1}{r}{1.0} & \multicolumn{1}{r}{0.14} & \multicolumn{1}{r}{1.0} & \multicolumn{1}{r}{1.0} & \multicolumn{1}{r}{0.38} \\

            \multicolumn{1}{l}{w/o CR} & \multicolumn{1}{l}{w/o ET} & \multicolumn{1}{r}{1.0} & \multicolumn{1}{r}{1.0} & \multicolumn{1}{r}{0.47} & \multicolumn{1}{r}{$0.010^*$} & \multicolumn{1}{r}{1.0} & \multicolumn{1}{r}{1.0} & \multicolumn{1}{r}{1.0} & \multicolumn{1}{r}{1.0} \\

            \multicolumn{1}{l}{w/o CR} & \multicolumn{1}{l}{w/o Hm} & \multicolumn{1}{r}{1.0} & \multicolumn{1}{r}{1.0} & \multicolumn{1}{r}{1.0} & \multicolumn{1}{r}{1.0} & \multicolumn{1}{r}{0.30} & \multicolumn{1}{r}{0.10} & \multicolumn{1}{r}{1.0} & \multicolumn{1}{r}{1.0} \\

            \multicolumn{1}{l}{w/o ET} & \multicolumn{1}{l}{w/o Hm} & \multicolumn{1}{r}{1.0} & \multicolumn{1}{r}{1.0} & \multicolumn{1}{r}{1.0} & \multicolumn{1}{r}{$0.010^*$} & \multicolumn{1}{r}{$0.030^*$} & \multicolumn{1}{r}{0.47} & \multicolumn{1}{r}{0.38} & \multicolumn{1}{r}{0.72} \\ \bottomrule
        \end{tabular}
        \vspace{1mm}
        \begin{tablenotes}
          \item[$\rm \dag{a}$]\footnotesize{is the abbreviation of ``Mental demand", which is one of NASA TLX subscales.}
          \item[$\rm \dag{b}$]\footnotesize{is the abbreviation of ``Physical demand", which is one of NASA TLX subscales.}
          \item[$\rm \dag{c}$]\footnotesize{is the abbreviation of ``Temporal demand", which is one of NASA TLX subscales.}
          \item[$\rm \dag{d}$]\footnotesize{is the abbreviation of ``Performance", which is one of NASA TLX subscales.}
          \item[$\rm \dag{e}$]\footnotesize{is the abbreviation of ``Effort", which is one of NASA TLX subscales.}
          \item[$\rm \dag{f}$]\footnotesize{is the abbreviation of ``Frustration", which is one of NASA TLX subscales.}
          \item[*]\footnotesize{The asterisked p-values indicate that there is a significant difference between the two methods at the significance level of $0.05$.}
        \end{tablenotes}
\end{threeparttable}
\end{table*}

\subsection{Collection Time}
\begin{table}[tb]
\begin{threeparttable}[tb]
    \small
    \centering
        \caption{\small{Collection time [s] taken for 100 images in each of three trials for all seven participants. Each element shows the mean $\pm$ standard deviation of the time and of the increase-decrease rate.}} \tablab{web-app-collect-time}
        {\tabcolsep = 1.2mm
        \begin{tabular}{p{12mm}p{3mm}p{3mm}p{3mm}p{10mm}} \toprule
            & \multicolumn{3}{c}{Collection time for each trial [s]} & \\ \cmidrule(r){2-4}
            \multicolumn{1}{c}{Method} & \multicolumn{1}{c}{1st} & \multicolumn{1}{c}{2nd} & \multicolumn{1}{c}{3rd} & \multicolumn{1}{c}{ID rate$^{\rm \dag{a}}$} \\ \midrule
            ALL & \multicolumn{1}{r}{159$\pm$47.4} & \multicolumn{1}{r}{176$\pm$51.7} & \multicolumn{1}{r}{213$\pm$71.9} & \multicolumn{1}{r}{0.400$\pm$0.516} \\
            w/o Hm & \multicolumn{1}{r}{78.0$\pm$23.4} & \multicolumn{1}{r}{74.0$\pm$14.6} & \multicolumn{1}{r}{76.1$\pm$20.3} & \multicolumn{1}{r}{-0.0117$\pm$0.133} \\
            w/o CR & \multicolumn{1}{r}{146$\pm$42.2} & \multicolumn{1}{r}{155$\pm$49.8} & \multicolumn{1}{r}{192$\pm$61.7} & \multicolumn{1}{r}{0.399$\pm$0.606} \\
            w/o ET & \multicolumn{1}{r}{173$\pm$27.5} & \multicolumn{1}{r}{167$\pm$39.9} & \multicolumn{1}{r}{264$\pm$114} & \multicolumn{1}{r}{0.487$\pm$0.406} \\ \bottomrule
        \end{tabular}
        }
        \vspace{1mm}
        \begin{tablenotes}
          \item[$\rm \dag{a}$]\footnotesize{shows the increase-decrease rate calculated by dividing the collection time for the final trial by the collection time for the first trial.}
        \end{tablenotes}
\end{threeparttable}
\end{table}
\tabref{web-app-collect-time} shows mean and standard deviation values of the time taken to collect 100 images in each of three trials for all seven participants. 
In addition, we calculated the increase-decrease rate (ID rate) obtained by dividing the collection time for the final trial by the collection time for the first trial.
\tabref{web-app-collect-time} further shows the mean and standard deviation values of the ID rate for all participants.
This value is related to the degree of familiarity of users with the collection operation and the cumulative workload as the number of trials increases.
A negative value indicates a presence of familiarity, whereas a positive value indicates a presence of fatigue.

In the case of the ALL mode, it took less than 10 [min] for data collection in the three trials, indicating that only 10 [min] was necessary to collect 300 images (100 images $\times$ 3 trials).
Compared with other methods, the mean value of the collection time by the ALL method was longer than that of the method without Hm, shorter than that by the method without ET, and almost the same as that of the method without CR. 
In the case of the method with Hm, because users tend to check whether the target rectangle disappears during mode execution carefully, the time was relatively long compared to that obtained by the method without Hm.

In the method without Hm, the collection rate was improved depending on changes in the pose of the camera view; thus, the collection rate tended to increase more easily.
Meanwhile, the time and ID rates for the three methods other than those without Hm were almost the same. The maximum difference in each trial was 37.5\% (72 [s]) for the shortest time.
In the method without ET, the time was longer in the 1st and 3rd trials compared to the corresponding times of the ALL method, and the ID rate was slightly higher.
The ID rate results showed the same tendency as the collection time results.

\subsection{Variation}
\begin{figure}[tb]
  \centering
  \begin{minipage}[tb]{0.90\linewidth}
    \centering
    \includegraphics[width=\linewidth]{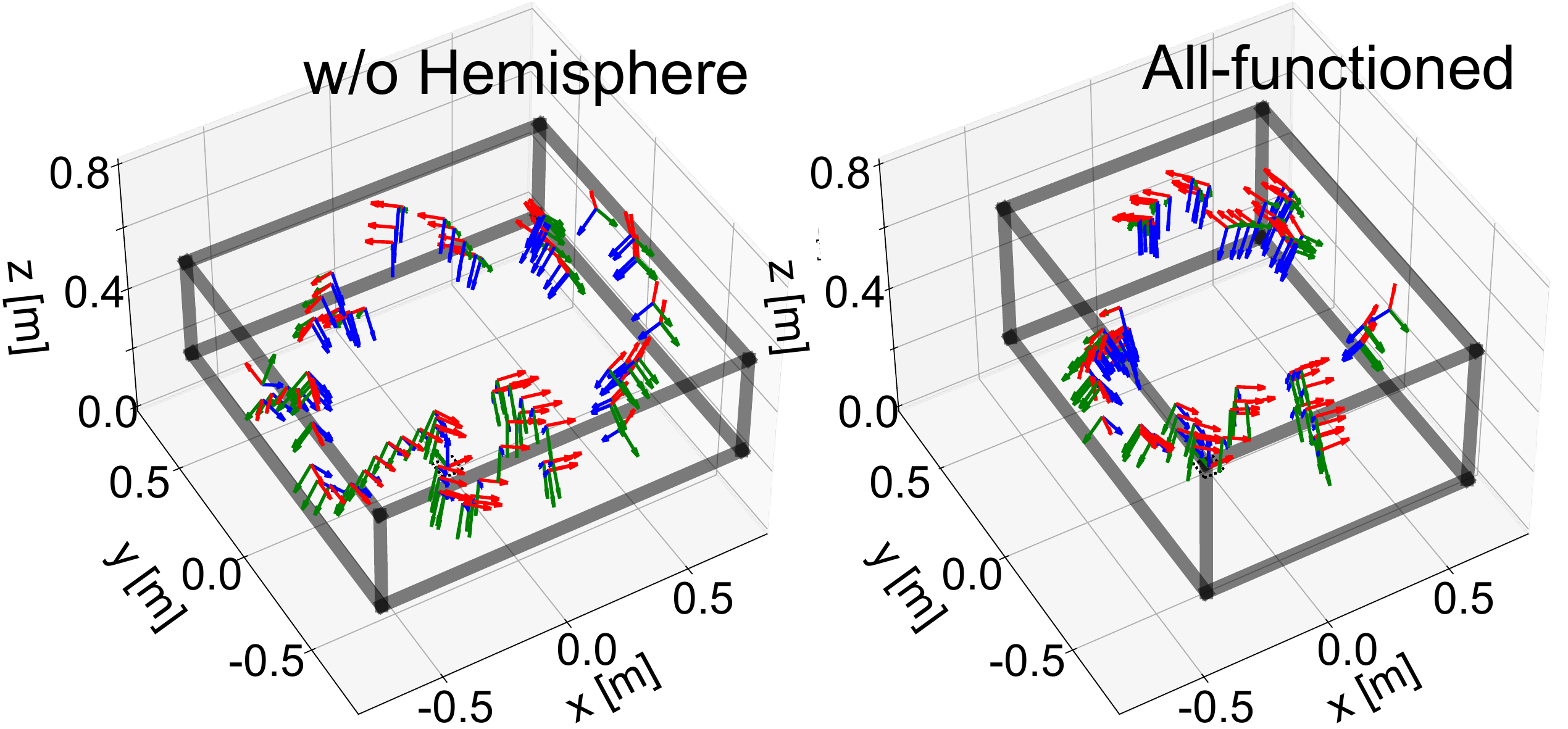}
    \subcaption{\small{First trial}}
  \end{minipage}
  \begin{minipage}[tb]{0.90\linewidth}
    \centering
    \includegraphics[width=\linewidth]{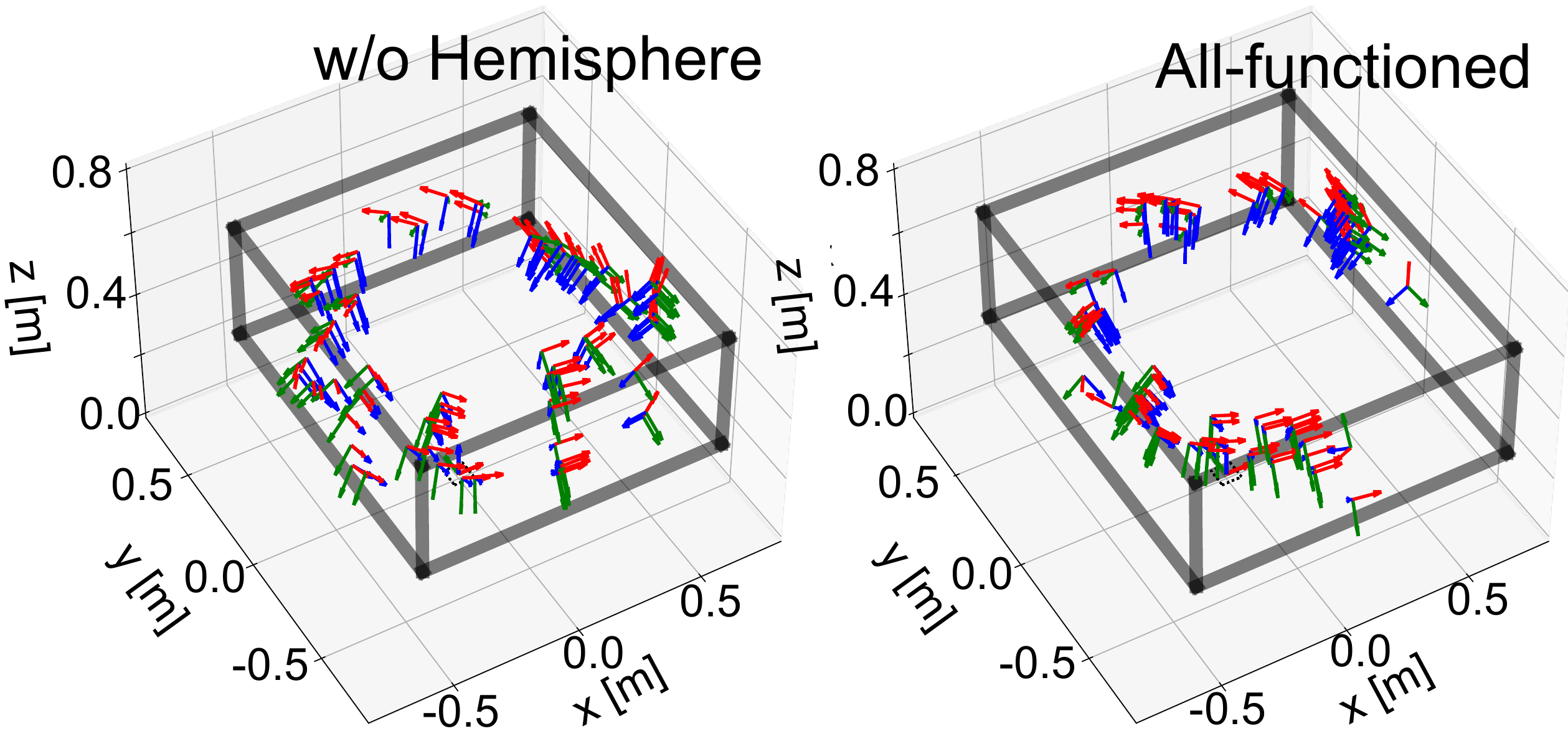}
    \subcaption{\small{Second trial}}
  \end{minipage}
  \begin{minipage}[tb]{0.90\linewidth}
    \centering
    \includegraphics[width=\linewidth]{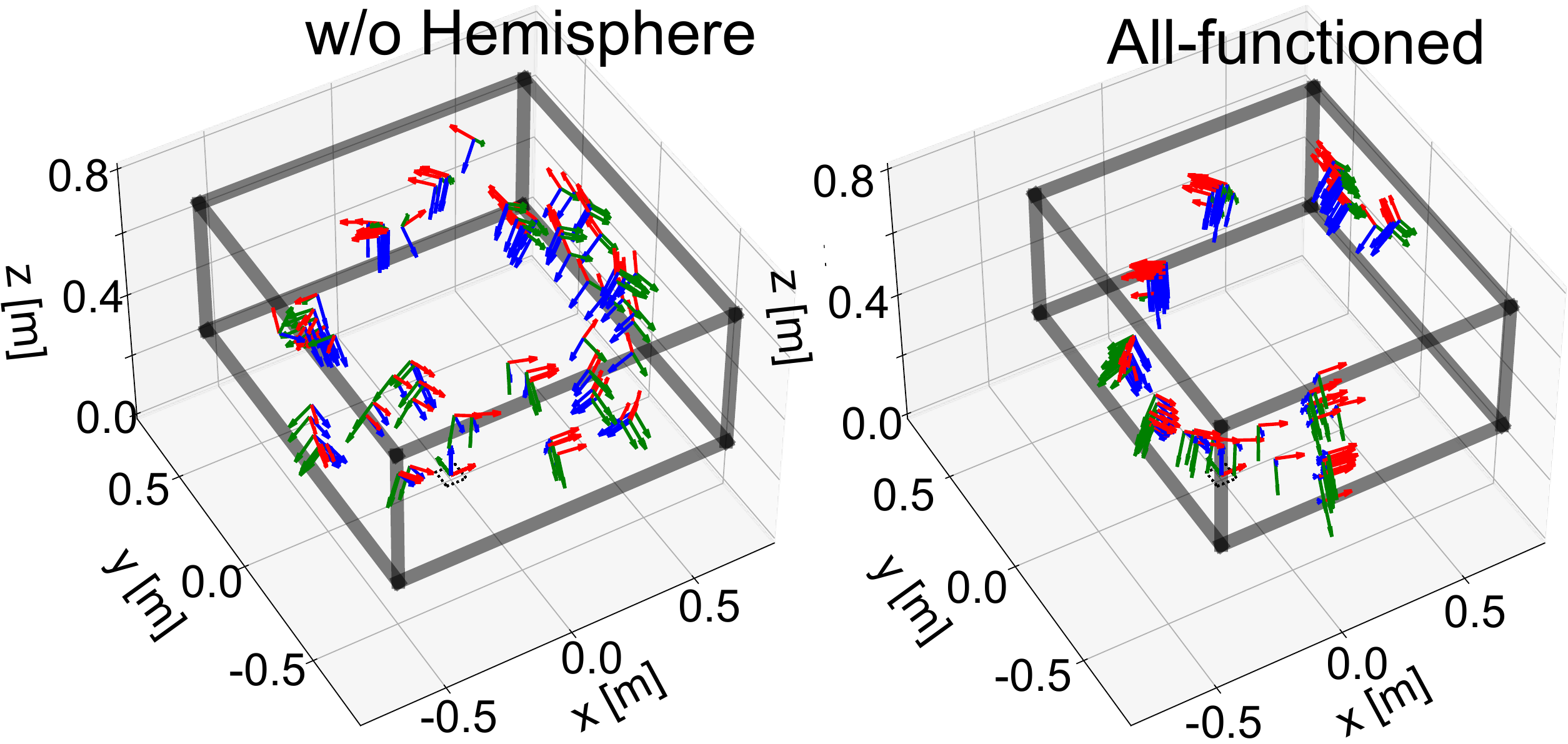}
    \subcaption{\small{Third trial}}
  \end{minipage}
  \caption{\small{Variation of collected datasets. The coordinate systems with the red, blue, and green arrows located at $z=0$ indicate the coordinate systems of the marker. The remaining coordinate systems show the camera poses where images were captured. The gray bounding box represents the smallest rectangular parallelepiped that contains all camera viewpoints.}}
  \figlab{result}
\end{figure}
\begin{table}[tb]
\begin{threeparttable}[tb]
    \small
    \centering
        \caption{\small{Calculated several types of variabilities in terms of distance [m] from the camera to object, the volume of 2D bounding box containing all positions, and angular distance [$^{\circ}$] in all viewpoints. For the two measures, this table shows the mean $\pm$ standard deviation values.}} \tablab{variation}
        {\tabcolsep = 2mm
        \begin{tabular}{p{12mm}p{15mm}p{15mm}p{15mm}} \toprule
            & \multicolumn{3}{c}{Distance for each trial [m]} \\ \cmidrule(r){2-4}
            \multicolumn{1}{c}{Method} & \multicolumn{1}{c}{1st} & \multicolumn{1}{c}{2nd} & \multicolumn{1}{c}{3rd} \\ \midrule
            ALL & \multicolumn{1}{r}{0.762$\pm$0.0759} & \multicolumn{1}{r}{0.733$\pm$0.0729} & \multicolumn{1}{r}{0.738$\pm$0.0866} \\ 
            w/o Hm & \multicolumn{1}{r}{0.678$\pm$0.133} & \multicolumn{1}{r}{0.691$\pm$0.132} & \multicolumn{1}{r}{0.729$\pm$0.134} \\ \midrule
            & \multicolumn{3}{c}{Volume [$\rm{m}^3$]} \\ \cmidrule(r){2-4}
            \multicolumn{1}{c}{Method} & \multicolumn{1}{c}{1st} & \multicolumn{1}{c}{2nd} & \multicolumn{1}{c}{3rd} \\ \midrule
            ALL & \multicolumn{1}{r}{0.595} & \multicolumn{1}{r}{0.534} & \multicolumn{1}{r}{0.747} \\
            w/o Hm & \multicolumn{1}{r}{0.631} & \multicolumn{1}{r}{0.611} & \multicolumn{1}{r}{0.542} \\ \midrule
            & \multicolumn{3}{c}{Angular distance for each trial [$^{\circ}$]} \\ \cmidrule(r){2-4}
            \multicolumn{1}{c}{Method} & \multicolumn{1}{c}{1st} & \multicolumn{1}{c}{2nd} & \multicolumn{1}{c}{3rd} \\ \midrule
            ALL & \multicolumn{1}{r}{152$\pm$16.8} & \multicolumn{1}{r}{156$\pm$19.7} & \multicolumn{1}{r}{157$\pm$14.7} \\
            w/o Hm & \multicolumn{1}{r}{158$\pm$16.5} & \multicolumn{1}{r}{155$\pm$17.6} & \multicolumn{1}{r}{155$\pm$20.4} \\
            \bottomrule
        \end{tabular}
        }
\end{threeparttable}
\end{table}
To reduce the number of uncaptured areas around a target object, it is important to collect multi-view image datasets.
To validate the effect of Hm on unbiased dataset collection, we compared the datasets generated by the ALL method with those generated by the method without Hm.
Note that in the method without Hm, the collection rate increases based on the pose of the camera view.

\figref{result} shows scatter plots of the camera's 6D poses used to capture the 100 images collected with the ALL method and the method without Hm. The figure also illustrates a 2D bounding box representing the smallest rectangular parallelepiped, containing all camera viewpoint positions (plotted points).
The wide distribution of the plotted points in the figure indicates that the variation in the dataset is large. 

Therefore, it appears that the variation in the dataset collected without Hm was also large. 
Nevertheless, if we observe the calculated distance [m], the ALL method shows larger differences in the distance for all trials compared with the method without Hm.
The two calculated volumes [$\mathrm{m}^3$] of the 2D bounding boxes containing each camera viewpoint position shown in~\tabref{variation} are similar because the mean distance in the method without Hm is larger.
In the ALL method, the coordinate systems showing the viewpoints appear relatively densely gathered.  These results are likely because the user carefully moved the camera incrementally to capture all visualized rectangles. Conversely, when the hemisphere was hidden, the camera moved randomly.
The angular distance is calculated as $2 \times \arccos(\bm{q^{*}} \cdot \bm{\hat{q}})$. $\bm{q^{*}}$ and $\bm{\hat{q}}$ represent the calculated quaternion, indicating the camera posture, and the unit quaternion, indicating the marker posture, respectively. The calculated angular distances were comparable between the ALL method and the method without Hm.

Therefore, the effect of hemisphere visualization allowed users to collect varied multi-view image datasets based on the distance between the camera and object positions.
However, regardless of whether Hm is visible, the variation does not change significantly because it is required to capture the scene from different viewpoints to increase the collection rate in all cases.

\section{Evaluating Annotation Results}
Note that this evaluation is not the main scope of this study.
Although we present an example of the results of the collected dataset as supplemental data, the results do not affect the effectiveness or lack thereof of the ALL method.

This section describes the annotation feasibility and detection performance of a vision system trained on a dataset of 500 images collected by a user, who also collected an additional 100 images for use as a test dataset.
The user is different from those who participated in the experiment described in~\textbf{Section~5}.
\figref{object} shows the six objects used in this evaluation.
We selected objects from the \textit{NEDO ITEM DATABASE}\footnote{http://mprg.cs.chubu.ac.jp/NEDO\_DB/}.
To evaluate the performance, we calculated the \textit{Intersection over Union} (IoU), \textit{precision}, \textit{recall}, and \textit{F-score} on the annotation and the \textit{average precisions} (APs) and \textit{average recalls} (ARs) for object detection.
\begin{figure}[tb]
  \centering
  \begin{minipage}[tb]{0.67\linewidth}
    \centering
    \includegraphics[width=\linewidth]{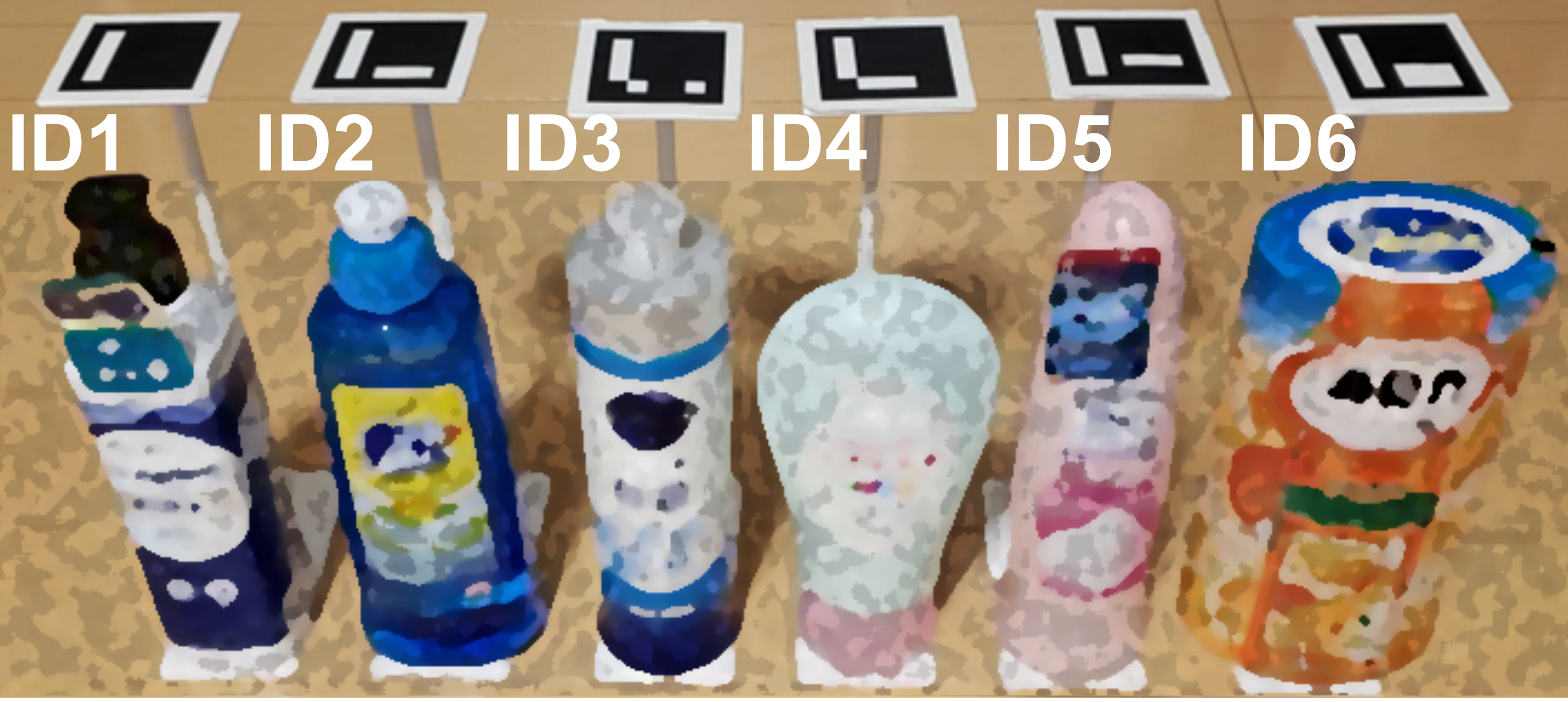}
    \subcaption{\small{Appearance}}
  \end{minipage}
  \begin{minipage}[tb]{0.29\linewidth}
    \centering
    \includegraphics[width=\linewidth]{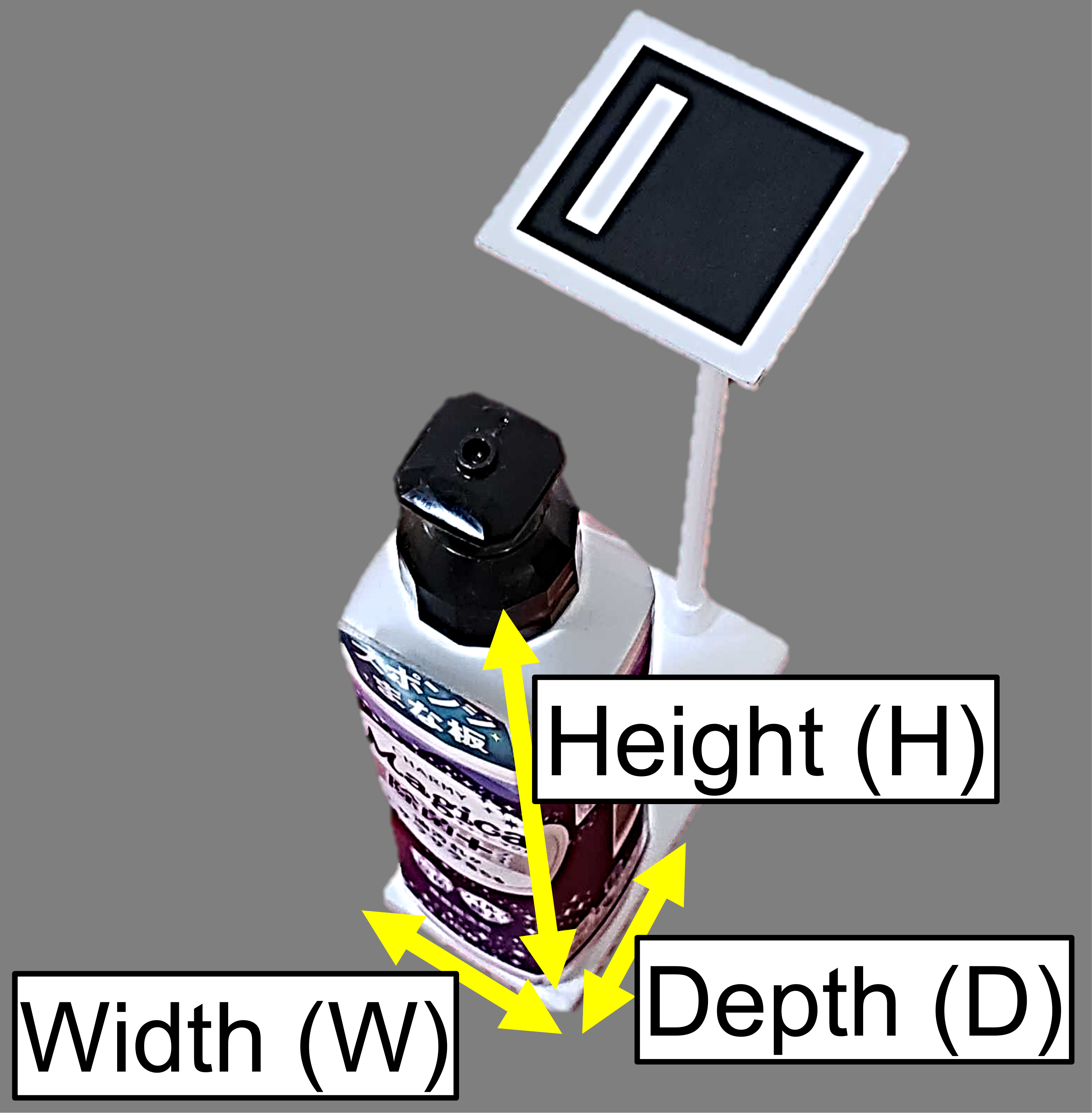}
    \subcaption{\small{Geometry}}
  \end{minipage}
  \caption{\small{Target objects for annotation feasibility evaluation.}}
  \figlab{object}
\end{figure}

\subsection{Annotation Feasibility}
\figref{annotation}\textbf{(b)} shows a visualization example of the annotation results.
To evaluate the annotation results of the ALL method using the four indicators, we compared the results with manually annotated images.
Using a manual annotation tool called \textit{labelme}\footnote{https://github.com/wkentaro/labelme}, a bounding box covering the object contours can be generated by humans.
The human annotators specified the top-left and bottom-right corners using a tool to create bounding boxes.
Manual annotation of 100 images required 28 [min] on average, whereas automatic annotation required less than 1 [s], indicating a significant improvement in the time performance.

By calculating the \textit{true-positive} (TP), \textit{false-positive} (FP), and \textit{false-negative} (FN) values, as shown in~\figref{anno-result}, we obtained the IoU, precision, recall, and F-score.
\tabref{data-seg-res} shows the results of object region extraction in the annotations.
The results show a type of similarity between the results of the proposed automatic annotations and manual annotations, and not the accuracy of the ALL method.
The table shows the calculated values of the mean $\pm$ standard deviation of the IoU, precision, recall, and F-score.

The mean values of the four indicators for all categories were 59.5\%, 60.5\%, 97.3\%, and 73.9\%, respectively.
The mean value of recall was rated as the highest, with smaller standard deviations than those of the other metrics. 
These results suggest that the same areas included in the manual annotations are rarely missed.
Moreover, the bounding boxes generated by the ALL method were larger than those generated by manual annotation.
In other words, there were a few over-annotated pixels in the proposed annotations.
The discrepancy in the segmentation pixels was due to the bounding box generation process, which was based on the size of the approximated shape. 
The visual markers were detected with 100\% accuracy; thus, the calculated values of precision, recall, and F-score in terms of \textit{detection}, but not \textit{segmentation}, were 100\%.
Therefore, as long as the user defines how to generate the bounding box according to the use case, annotation works according to the user's preference.

\begin{figure}[tb]
  \centering
  \begin{minipage}[tb]{0.58\linewidth}
    \centering
    \includegraphics[width=\linewidth]{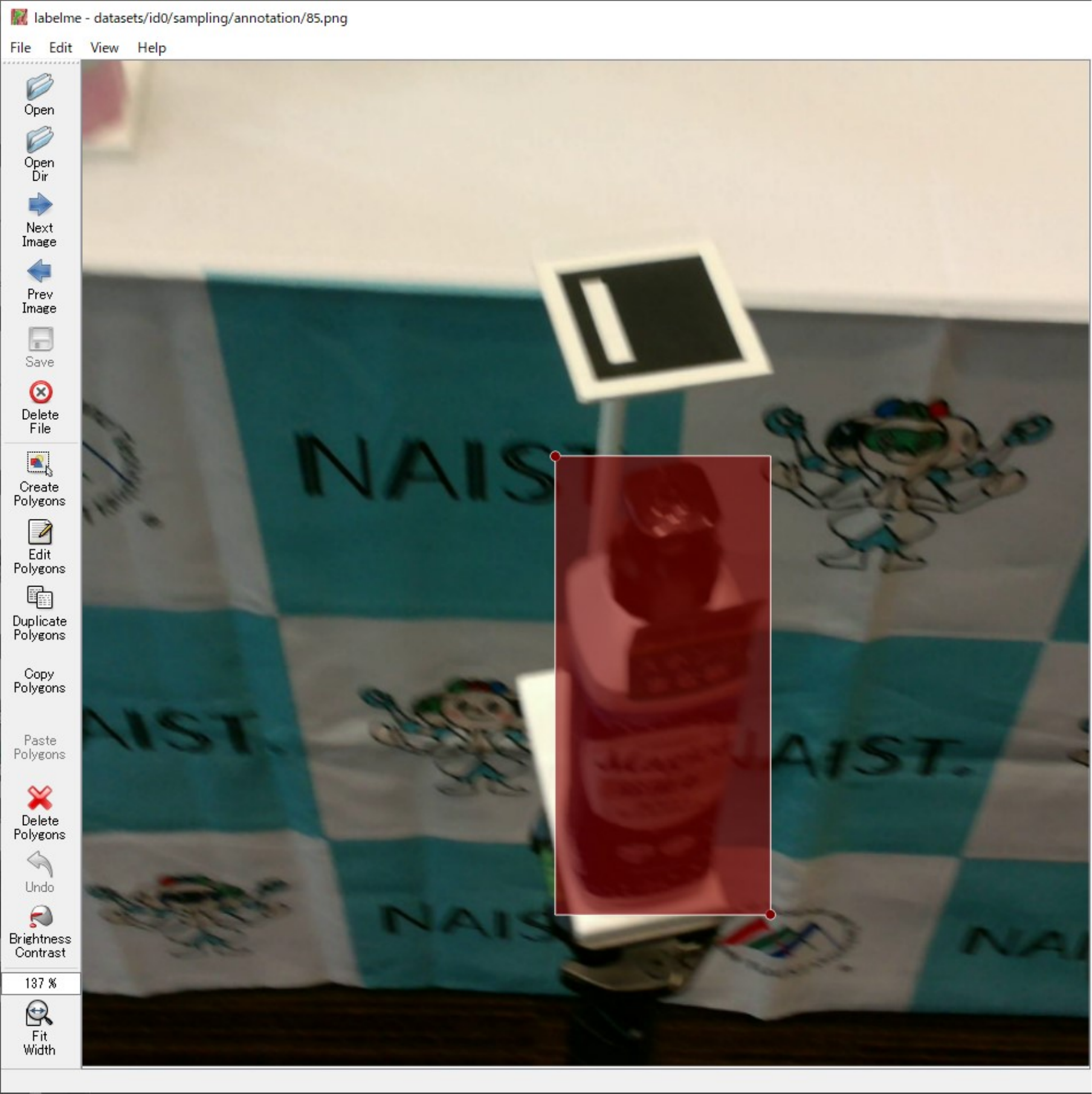}
    \subcaption{\small{Clicked points}}
  \end{minipage}
  \begin{minipage}[tb]{0.40\linewidth}
    \centering
    \includegraphics[width=\linewidth]{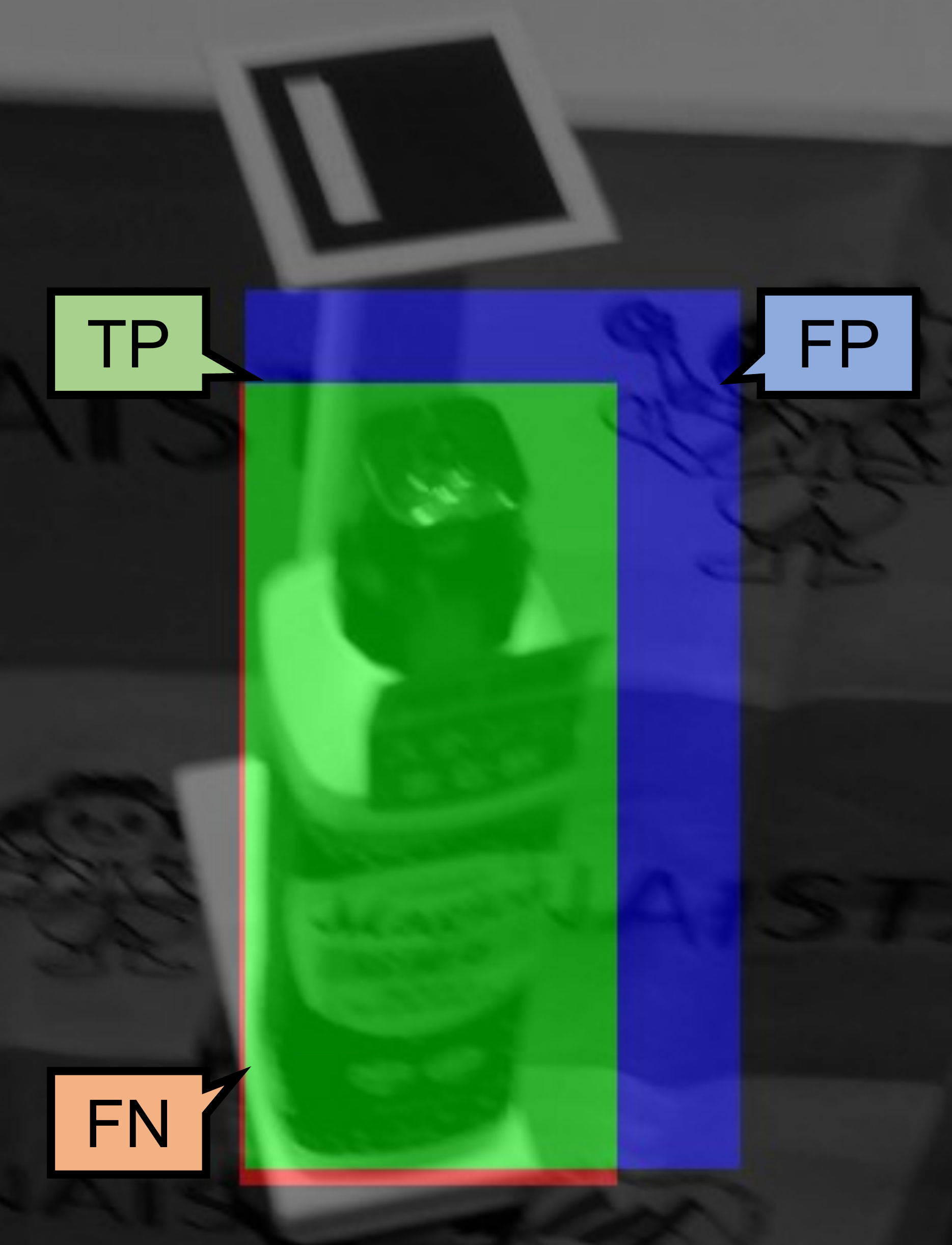}
    \subcaption{\small{Class of image region}}
  \end{minipage}
  \caption{\small{Manual annotation. The left picture shows an example annotation with a tool (labelme). The right illustration shows the classes assigned to image regions, which are used for evaluating the region extraction.}}
  \figlab{anno-result}
\end{figure}
\begin{table}[tb]
    \small
    \centering
    \caption{\small{Performance of annotations. Each element shows mean $\pm$ standard deviation of IoU [\%], Precision [\%], Recall [\%], and F-score [\%]. The values represent the similarity of the two values, not the accuracy.}}
    {\tabcolsep = 1.4mm
    \begin{tabular}{p{6mm}p{16mm}p{16mm}p{16mm}p{16mm}} \toprule
        & \multicolumn{4}{c}{{Metric [\%]}} \\ \cmidrule(r){2-5}
        ID$^{\rm \dag{a}}$ & \hfil{IoU}\hfil & \hfil{Precision}\hfil & \hfil{Recall}\hfil & {F-score} \\ \midrule
        1 & \multicolumn{1}{r}{68.9$\pm$7.19} & \multicolumn{1}{r}{70.4$\pm$7.92} & \multicolumn{1}{r}{97.6$\pm$3.97} & \multicolumn{1}{r}{81.4$\pm$5.09} \\
        2 & \multicolumn{1}{r}{62.2$\pm$10.3} & \multicolumn{1}{r}{63.7$\pm$10.1} & \multicolumn{1}{r}{95.7$\pm$10.8} & \multicolumn{1}{r}{76.1$\pm$9.60} \\ 
        3 & \multicolumn{1}{r}{58.7$\pm$9.83} & \multicolumn{1}{r}{59.1$\pm$10.2} & \multicolumn{1}{r}{99.2$\pm$1.52} & \multicolumn{1}{r}{73.5$\pm$7.67} \\
        4 & \multicolumn{1}{r}{49.3$\pm$11.9} & \multicolumn{1}{r}{50.5$\pm$13.6} & \multicolumn{1}{r}{96.5$\pm$11.1} & \multicolumn{1}{r}{65.2$\pm$11.4} \\
        5 & \multicolumn{1}{r}{61.3$\pm$10.9} & \multicolumn{1}{r}{62.5$\pm$11.1} & \multicolumn{1}{r}{96.0$\pm$10.1} & \multicolumn{1}{r}{75.3$\pm$10.2} \\
        6 & \multicolumn{1}{r}{56.4$\pm$9.66} & \multicolumn{1}{r}{56.9$\pm$10.5} & \multicolumn{1}{r}{98.9$\pm$2.18} & \multicolumn{1}{r}{71.7$\pm$7.39} \\ \midrule
        Mean &  \multicolumn{1}{r}{59.5$\pm$5.96} & \multicolumn{1}{r}{60.5$\pm$6.15} & \multicolumn{1}{r}{97.3$\pm$1.36} & \multicolumn{1}{r}{73.9$\pm$4.89} \\
        \bottomrule
    \end{tabular}
    }
    \vspace{1mm}
    \begin{tablenotes}
        \item[${\rm \dag{a}}$]\footnotesize{$^{\rm \dag{a}}$~The number of ID corresponds to the number shown in~\figref{object}.}
    \end{tablenotes}
    \tablab{data-seg-res}
\end{table}

\subsection{Detection Performance}
This experiment evaluates a vision system trained using an image dataset collected using the ALL method.
To evaluate the quality of annotation, we compared the developed automatic annotation with manual annotation under the same conditions.
We used EfficientDet D1~\cite{Tan2020}, based on a deep neural network model, as the object detector and trained the model using the collected dataset.
We confirmed the detection results of the two models trained using the proposed and manual datasets.
The APs (50\% $<$ IoU $<$ 95\%) were 74\% and 78\% for the proposed and manual methods, respectively.
The ARs (50\% $<$ IoU $<$ 95\%) for the proposed and manual methods were 78\% and 81\%, respectively.
These results demonstrate that the accuracies of the two models are almost the same.
In this experiment, although we collected only 100 images for each object image dataset if we could collect more datasets, we could use another EfficientDet model with higher compound coefficients (D2, D3, D4, D5, or D6), and the results might be improved.

As the same performance as that of the manually collected dataset was obtained, the effectiveness of the proposed web application did not decrease.

\section{Discussion}
\begin{figure}[tb]
  \centering
  \includegraphics[width=\linewidth]{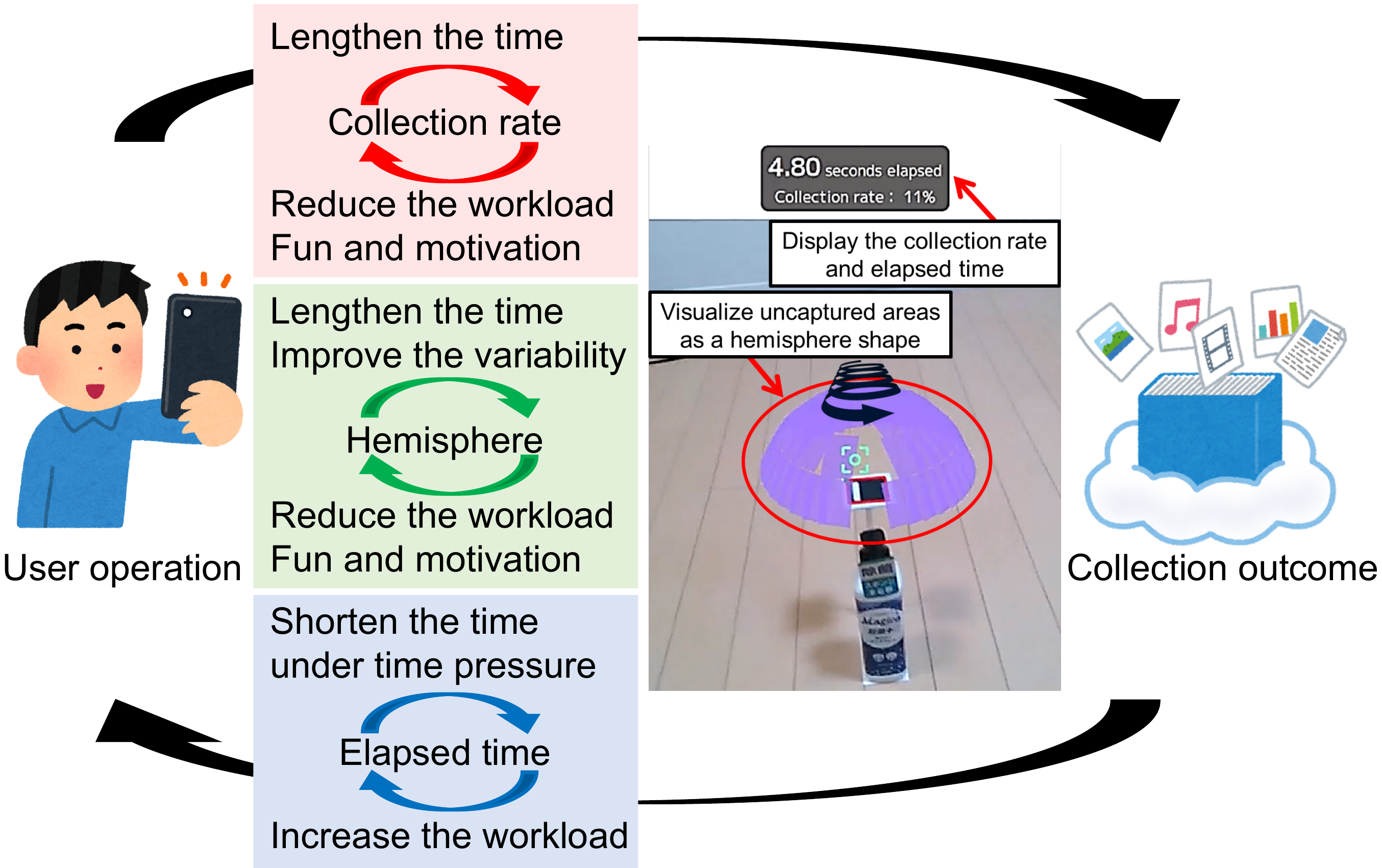}
  \caption{\small{Relationship among the application functions that compensate for the disadvantages of each other.}}
  \figlab{relation}
\end{figure}
\subsection{Motivation from Enjoyment, Efficiency, and Variation in Dataset Collection}
As indicated by the results described in~\textbf{Section~5.1}, the ALL method and the method without ET showed low scores (better outcome) for temporal demand.
In addition, the method without Hm had the lowest self-evaluation of performance by the user, indicating that the case using the method with Hm tends to rate the quality of the trials higher.
As described in~\textbf{Section~5.3}, the method with Hm might have an effect of giving motivation from enjoyment in their dataset collection trials.
In terms of efficiency, as explained in~\textbf{Section~5.4}, the method without Hm had the shortest collection time; however, its self-evaluation of performance was low, indicating that it might tend to estimate its performance as higher if it took the appropriate amount of time to collect the datasets.
Moreover, even for the dataset variation mentioned in~\textbf{Section~5.5}, it is preferable to show the hemispheres.

Considering these results as a whole, displaying the collection status such as Hm and CR might result in a time-consuming collection trial. Nevertheless, combining the visualizing Hm, CR, and ET, as in the ALL method, makes the user feel enjoyment and allows them to concentrate more on their trials rather than feeling excessive time pressure; thus, the users evaluated the quality of their trials as more high-quality.
Because variation was better in the case of the method with Hm, the quality of their trials was higher, and they might have performed the collection trials with an awareness of the progress as positive feedback on the collection status.

In our proposed system, updates of CR and Hm during a collection trial were positive factors indicating progress toward the goal, whereas the increase of ET was a negative indicator signifying a decline in performance. This was possibly one of the factors that ET led to high load. To balance pressure and performance, it is necessary to redesign ET to have a positive impact. For example, using metrics such as collection efficiency (items/second) should serve as a reward for the user.

In summary, although the collection time could not be reduced considerably compared with other methods, the proposed application can motivate users in an enjoyable manner, and users can collect high-quality datasets in terms of accuracy and variation with a moderate workload such as time pressure to boost the collection.
\figref{relation} shows the discovered relationship among the application features that compensate for the disadvantages of each other.
Because even participants who had no experience using AR technology or did not know about training dataset collection tended to score results similar to those of other participants, participants without the knowledge benefited from online visual feedback and could efficiently collect datasets. We believe that the reliability and generality of the experimental results are not low.

\subsection{Highly Varied Multi-view Dataset Collection}
There is room for improvement in terms of the variation of the datasets collected.
In the experiments, since the number of images to be collected (can be set in \textit{Maximum number of images} shown in~\figref{transition}) was set to 100, the user captured the target object in different directions to change the camera view as much as possible instead of changing capturing distances.
No specific instructions were given to users.

In the future, if it is desired to collect datasets that vary more, although we can set a larger number appropriately, the collection time will be longer, and the workload will increase accordingly.
To address this issue, it might be effective to implement another modality function, such as a voice announcement function to report the elapsed time and a sound effect function to report the collection progress. In future studies, including the collection of a larger number of datasets, we will clarify whether such multi-modal online visual feedback techniques can further improve enjoyment for motivation.

\subsection{Precise Annotation}
In the experiments, participants used pre-calibrated cameras. However, for general users of the web application, calibration using markers or checkerboards may be difficult. Therefore, it is worth considering the use of camera self-calibration methods~\cite{Hemayed2003}, which estimate the camera’s intrinsic parameters without markers or checkerboards.

For example, COLMAP~\cite{Schonberger2016} is a structure from a motion application that has camera self-calibration options and can be easily used for the estimation of the camera’s intrinsic parameters using only multiple images.

\section{Conclusions}
This study presented a web application-based dataset collection system to avoid a laborious and monotonous annotation process.
We proposed displaying three types of dataset collection statuses on the display of the application, including the uncaptured area as a hemisphere, collection rate, and elapsed time, to collect multi-view image datasets with less mental workload and time pressure in an enjoyable manner to boost motivation.

The results of our user testing revealed both the beneficial and detrimental consequences of the proposed features.
In summary, as illustrated in~\figref{relation}, although displaying the hemisphere and collection rate tends to lengthen the collection time, the combination of these functions has the potential to reduce mental workload and allow users to enjoy it.
By displaying the elapsed time, the collection time can be shortened by realizing dataset collection under time pressure.
However, the mental workload caused by time pressure can be reduced by providing users with enjoyable and motivated operations using a display of the hemisphere and collection rate.
Furthermore, we discovered that the hemisphere display increased the variation in the dataset and provided a sense of satisfaction by estimating their performance to be higher.
These results indicate that the combination of the proposed online visual feedback functions may improve motivation and balance the user workload and performance.
These findings do not deviate significantly from the hypotheses presented in~\tabref{pros-cons}.

Furthermore, our experiments using the developed dataset collection system suggest the feasibility of annotation and object detection.
Our future studies will promote the development of a simple yet effective multi-modal online visual feedback application capable of efficiently collecting highly varied multi-view image datasets for several types of target objects.

\bibliographystyle{IEEEtran}
\footnotesize
\bibliography{reference}

\end{document}